\documentclass[12pt]{article}

\usepackage{arxiv}

\usepackage{microtype}
\usepackage{graphicx}
\usepackage{subcaption}
\usepackage{booktabs} % for professional tables
\usepackage{multirow}
\usepackage{natbib}
\usepackage{amsthm}
\usepackage{enumitem}
\usepackage{hyperref}
\hypersetup{
	colorlinks=true,       
	linkcolor=blue,         
	citecolor=blue,        
	filecolor=magenta,      
	urlcolor=cyan           
}

\theoremstyle{plain}
\newtheorem{theorem}{Theorem}
\newtheorem{proposition}[theorem]{Proposition}
\newtheorem{lemma}[theorem]{Lemma}
\newtheorem{corollary}[theorem]{Corollary}
\theoremstyle{definition}

\theoremstyle{remark}

\usepackage{amsmath}
\usepackage{amssymb}
\usepackage{mathtools}
\usepackage{amsthm}

\usepackage{algorithm}
\usepackage{algorithmic}

\usepackage[capitalize,noabbrev]{cleveref}

\newcommand{\tobs}{{\scalebox{0.5}{$\mathrm{obs}$}}}
\newcommand{\ttrain}{{\scalebox{0.5}{$\mathrm{train}$}}}
\newcommand{\ttest}{{\scalebox{0.5}{$\mathrm{test}$}}}
\newcommand{\tmin}{{\scalebox{0.5}{$\mathrm{min}$}}}
\newcommand{\tmaj}{{\scalebox{0.5}{$\mathrm{maj}$}}}
\newcommand{\ttru}{{\scalebox{0.5}{$\mathrm{tru}$}}}

\newcommand{\dataset}{\mathcal{D}}
\newcommand{\odataset}{\dataset^\tobs}
\newcommand{\minset}{\dataset^\tmin}
\newcommand{\majset}{\dataset^\tmaj}
\newcommand{\traindata}{\dataset^\ttrain}
\newcommand{\testdata}{\dataset^\ttest}
\newcommand{\strata}{\mathcal{S}}
\newcommand{\strat}{S}
\newcommand{\featspace}{\mathcal{X}}

\newcommand{\prox}{\eta}

\newcommand{\logit}{\mathrm{lgt}}
\newcommand{\numdim}{d}
\newcommand{\bdist}{\mathcal{P}}
\newcommand{\dist}{\bdist^\ttru}
\newcommand{\odist}{\bdist^\tobs}
\newcommand{\numinstance}{N}
\newcommand{\nummaj}{n^-}
\newcommand{\nummin}{n^+}
\newcommand{\obsindic}{O}
\newcommand{\obsprb}{\pi}
\newcommand{\prob}{P}

\newcommand{\classifier}{f}
\newcommand{\node}{m}
\newcommand{\child}{{v}}
\newcommand{\bin}{b}
\newcommand{\binset}{\mathcal{B}}
\newcommand{\quant}{q}
\newcommand{\const}{c}

\title{Learning under Localized Minority Imbalance}

\author {
    Amin Hosseininasab,
    Steven M. Shugan\\
    Warrington College of Business\\
    University of Florida\\
    Gainesville, FL 32611 USA\\
    \texttt{a.hosseininasab@ufl.edu, sms@ufl.edu}
}

\begin{document}

\maketitle

\begin{abstract}
  Class-imbalance methods implicitly assume that the minority class is uniformly undersampled relative to the majority class. However, in many real-world settings, minority instances may be disproportionately under-observed in certain regions of the feature space. For example, small businesses that go bankrupt may disappear from records, while those that survive remain visible, making bankruptcy appear less common among small firms than it actually is. This gives rise to localized minority imbalance (LMI), a challenge that is often overlooked and extends beyond general class-count imbalance. We show that under LMI, existing imbalance mitigation techniques can fit observation-induced biases in the training data and generalize poorly to under-observed regions of the true minority distribution. To address this, we propose a tree-based stratified approach that recursively partitions the feature space with the goal of reducing within-stratum LMI distortion. For each resulting stratum, we pair its majority instances with the full observed minority set and train a base classifier to create an ensemble. Extensive experiments over benchmark tabular datasets simulated with LMI show that our stratified ensembling approach outperforms popular and state-of-the-art imbalance mitigation techniques. We also introduce a gold-standard evaluation protocol that uses unbiased test sets, and demonstrate that conventional hold-out evaluation from the same LMI-biased data can substantially mislead performance. Overall, our results highlight that the cause of imbalance is as important as the correction method.
\end{abstract}

\section{Introduction}

Class imbalance is prevalent in real-world applications and can substantially affect supervised learning performance. When the distribution of classes is highly uneven, algorithms learn to prioritize the majority class and become poor detectors of the minority class. This is problematic, especially in social sciences and applications where the minority class is the primary focus of inference. Examples are customer churn, purchase conversion, fraud detection, and medical diagnosis.

Most existing imbalance methods inherently assume that the relative observation propensity of the two classes is uniform across the feature space. Under this condition, imbalance induces a constant shift in the observed class log-odds, which can be effectively mitigated using existing techniques. However, in practice, the relative observation propensity may vary across feature values, a condition that we term \emph{Localized Minority Imbalance} (LMI).

\begin{figure*}
     \centering
     \subfloat[][\centering Balanced Dataset.]{
         \centering
         \includegraphics[scale=0.2]{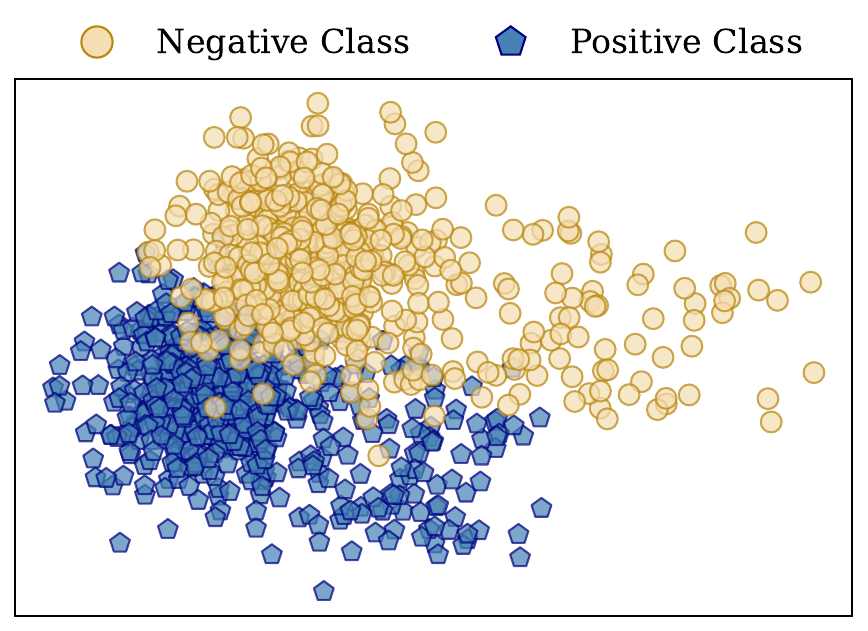}
         \label{fig:ex_bal}
     }
     \subfloat[][\centering Class-count Imbalance.]{
         \centering
         \includegraphics[scale=0.2]{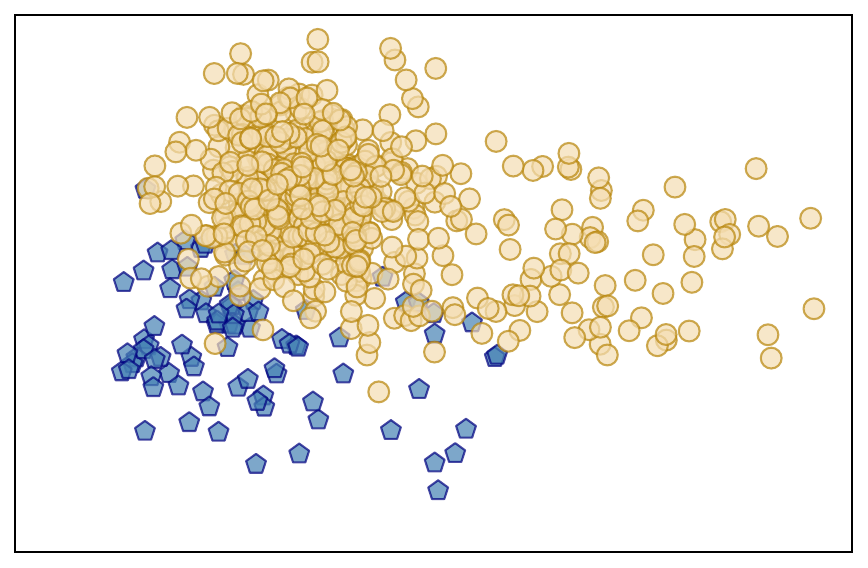}
         \label{fig:ex_mar}
     }
    \subfloat[][\centering LMI.]{
         \centering
         \includegraphics[scale=0.2]{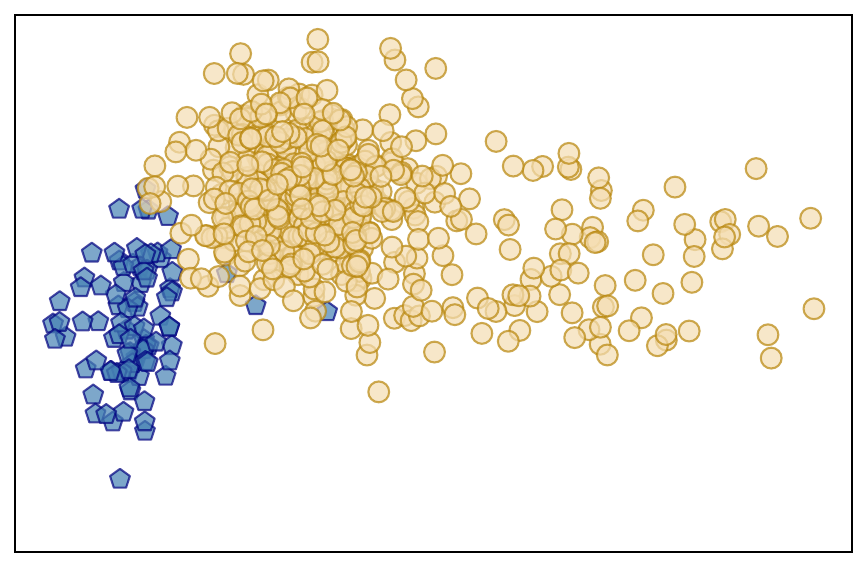}
         \label{fig:ex_mnar}
     }
    \caption{A balanced dataset with corresponding class-count and LMI scenarios. Both imbalanced datasets have the same number of instances, but positive instances with higher x-axis values are less likely observed under LMI, making them less representative of their true distribution.}
    \label{fig:example_imbl}
\end{figure*}

Under LMI, certain regions of the minority class are disproportionately unobserved---beyond what their natural frequency in the data-generating distribution would suggest. This imbalance occurs precisely due to their class and feature values, resulting in selection bias, covariate shift, or similar paradigms. For example, the bankrupt firms we observe (minority) may not represent all bankruptcies, \textit{since} small failed firms often vanish from records. This is unlike small non-bankrupt firms which remain visible. Similarly, a demographic with severe depression (minority) may be less likely than other demographics to respond to questions about emotional well-being, particularly due to their level of mental health.

LMI can arise whenever the process that reveals, records, labels, or retains positive outcomes varies across feature-defined subpopulations. Examples include differential reporting, diagnosis or detection, survey response, record retention, and outcome ascertainment. Importantly, because the true posterior and the feature-dependent observation mechanism are not separately identifiable from the observed data alone (\S4.1), the empirical prevalence of LMI cannot generally be estimated without external information. We therefore do not claim that LMI is universal. Rather, the framework is intended for applications in which the data-collection process provides a substantive reason to suspect that minority observability differs across feature regions.

The scarcity of minority observations can make the consequences of LMI particularly severe. Smaller samples of the population have a higher risk of underrepresenting or even excluding less common subpopulations. As demonstrated in Figure~\ref{fig:example_imbl}, this can lead to a biased representation of the minority population, and adds an additional challenge on the already difficult class-count imbalance problem. Our work exposes this overlooked bias and reframes imbalance learning through the lens of LMI.

We first formalize this seemingly overlooked type of imbalance, and show that under LMI, existing imbalance mitigation techniques can reinforce the very biases they aim to correct. In particular, they tend to overfit the majority class in regions where the minority population is less-observed, and therefore, generalize poorly to the true population. 

We then show that LMI distortion can be mitigated by a careful selection and pairing of majority subsets from under-observed minority regions. Training over the entire observed dataset leads to abundant evidence for the majority class in parts of the feature space where minority instances truly occur but are less-observed. As a result, majority points distort classifiers towards predicting their class. However, by restricting the majority class to a carefully chosen subset according to LMI regions, we can reduce precisely the source of distortion that exists over the entire dataset.

Using this as motivation, we develop a novel tree-based procedure that seeks to generate strata with reduced within-stratum LMI distortion. Our Stratified Ensembling algorithm (SE) is guided by a custom objective function that uses observable parent-child prevalence changes as a surrogate for LMI distortion and recursively partitions the feature space while this objective improves. The procedure therefore seeks strata with reduced within-stratum LMI distortion, without requiring identification of the underlying observation mechanism. This then allows us to train a local classifier on each stratum, and apply existing imbalance mitigation techniques within the resulting ensemble. 

While stratifying the feature space leads to lower LMI severity in each stratum, it also introduces higher variance due to a smaller and region-specific sample of data. This adversely affects the generalization of each local classifier to the full data distribution. For an effective bias-variance tradeoff, we propose to (1) combine the majority instances of each stratum with the entire set of minority instances, and (2) ensemble all classifiers for the final prediction task. 

First, pairing all minority instances increases the effective sample size of the scarce class in every base classifier, leading to decision boundaries that are less sensitive to region-specific minorities. This is especially important under LMI, where local minority subsets can be rare and highly idiosyncratic. Second, restricting the majority class to each stratum preserves the intended localization---each local classifier sees a simpler, more homogeneous slice of the majority distribution---mitigating the imbalance effect of the majority class in strata with under-observed minority instances. Lastly, ensembling across strata further lowers variance by averaging out stratum-specific estimation noise.

We compare our approach to popular and state-of-the-art class imbalance techniques over several base classifiers and benchmark datasets, and show that under LMI our approach achieves higher MCC, F1 on unbiased test sets. On the other hand, when imbalance is class-count, existing methods perform the best. SE is thus most appropriate when the data-collection process provides a substantive reason to expect feature-dependent minority observability. When a less-biased validation sample is available, SE and standard methods can be compared directly on that sample, while in its absence, SE should be interpreted as a robustness analysis rather than as a data-only diagnosis of LMI.

Lastly, we address an issue in evaluating imbalance methods under LMI: standard hold-out splits drawn from LMI datasets are biased and mislead performance assessments. Specifically, standard hold-out sampling evaluates performance primarily on the observed minority regions rather than the true target population. Therefore, evaluating imbalance mitigation techniques without considering the source of imbalance can lead to misleading conclusions, and even favor methods that overfit the selection pattern instead of learning the true data distribution. 

To address this, we propose a gold-standard validation protocol that requires the test set to be LMI-free. While such balanced reference test sets may not always be readily available, they can be constructed when LMI is suspected in many realistic settings. For example, researchers can collect targeted labels, use external cohorts, follow up on delayed outcomes, or audit data-collection procedures to balance observation in LMI-suspected regions. Nonetheless, our goal in this paper is not to propose a universally adoptable evaluation standard. Rather, we seek to highlight the potential pitfalls of conventional evaluation practices when class imbalance arises from LMI, and showcase how our evaluation protocol can reveal improvements that translate to better coverage of the minority subgroups suppressed by LMI. Our contributions can be summarized into four parts: 
\begin{itemize}
    \item We formalize LMI and provide a theoretical analysis showing how partitioning the feature space into strata that approximate class-count imbalance can help mitigate its effects.
    \item We develop a novel SE procedure as a first step toward addressing LMI. We emphasize that our approach provides an approximate strategy for mitigating LMI, and also do not claim to improve over existing methods for general class-count imbalance.
    \item We empirically show that SE outperforms popular and state-of-the-art imbalance mitigation techniques, which while suited to class-count imbalance, fail to generalize under LMI.
    \item We show that standard hold-out sampling from LMI data misleads performance, and introduce an evaluation protocol that uses unbiased test sets for LMI correction. 
\end{itemize}
 
\section{Related Work}

A broad category of imbalance techniques falls into data-level, cost-sensitive, and hybrid strategies (see \citet{altalhan2025imbalanced} for a recent review). Data-level methods include oversampling the minority class, such as SMOTE \citep{chawla2002smote} and its variants \citep{pradipta2021smote}, which can mitigate imbalance but risk reinforcing minority class bias under LMI. On the other hand, undersampling methods like Cluster Centroids \citep{yen2009cluster} may lose important structure that can aid in mitigating LMI. Cost-sensitive learning modifies the weight of instances in classification loss functions to emphasize the minority class during training, such as focal loss \citep{lin2017focal}, class-balanced loss \citep{cui2019class}, and meta reweighting \citep{ren2018learning,shu2019meta}, but face similar limitations to oversampling under LMI. 

Ensemble and hybrid methods combine multiple techniques to leverage their strengths, and have gained popularity in recent years (see \citet{khan2024review} for a review). However, most ensembles rely on random undersampling, are designed mainly to reduce variance, and do not explicitly address LMI. In contrast, our method specifically promotes diversity across the ensemble to learn informative contrasts between majority and minority instances in under-observed minority regions.

Other imbalance methods focus on establishing rigorous guarantees for metrics relevant under class imbalance (e.g., MCC, F1) \citep{diochnos2021learning,narasimhan2018learning} or AUC-PR directives \citep{eban2017scalable,qi2021stochastic}. Similarly, other work addresses performance disparities between hard and easy classes through margin regularization \citep{zhu2026reducing}. While these approaches target metric optimization or class-wise generalization disparities, they do not directly addresses the feature-dependent observation bias underlying LMI. Our approach is complementary and can be applied in conjunction.

Another related stream is when the distribution of the test set is considered to \emph{shift} from the training set. Such shifts are studied in the context of class imbalance with both covariate and label shifts \citep{tan2020class,lopez2014importance,yu2025towards}, domain adaptation \citep{lee2024domain,fu2025rdam}, and concept drift \citep{palli2024online}. These works focus on adaptation, reweighting, or tracking changes over time, often using external information and access to unlabeled target-domain data. In contrast, we assume a single underlying population but a biased observation mechanism, no prior knowledge on the underlying distribution, and no access to auxiliary unlabeled data. As the training minority class-conditional is itself distorted in our framework, shift/adaptation methods do not address the core challenge of LMI.

Similarly, distributionally robust optimization \citep{namkoong2016stochastic,sagawa2019distributionally} has gained attention in training robust classifiers that extend to unseen environments. While these methods also seek robustness to distributional differences, they typically optimize against worst-case shifts over an uncertainty set. In contrast, our work targets a specific LMI bias that is feature-dependent and aims to correct that structured under-observation rather than generic distribution shift.

A related literature examines long-tailed classification, where class frequencies are highly unequal. The literature can be categorized into re-balancing \citep{ren2020balanced,kang2019decoupling,bhat2025prior2posterior}, logit and prior correction \citep{menon2020long}, and decoupled or multi-expert training \citep{kang2019decoupling,wang2020long}. The core goal of this stream is better tail generalization without sacrificing head performance. While these methods address predictive degradation under imbalance, they typically assume imbalance is global and tail classes are \emph{uniformly} underrepresented across the feature space (that is, class-count imbalance). Therefore, these methods mainly correct frequency-driven optimization and calibration, not the LMI distortion that we study. Our work is thus complementary, and particularly relevant when long tails are LMI.

\section{Problem Formalization} \label{sec:formal}

We consider a supervised classification task over an \emph{observed} dataset $\odataset = \left\lbrace \left( x_i, y_i \right) \right\rbrace_{i=1}^\numinstance$. Features $x_i = \left[ x^1_i,\dots,x^d_i \right]$ are defined over a $d$-dimensional feature space $X \in \featspace = \mathbb{R}^d$, and labels are binary $y_i \in \{0, 1\}$ (our method can be extended to multiclass via one-vs-rest). The observed dataset consists of $\nummaj$ observations from the $Y=0$ negative or \emph{majority} class, and $\nummin$ observations from the $Y = 1$ positive or \emph{minority} class. In many real-world applications, $\nummin \ll \nummaj$, creating class imbalance.

The observed dataset $\odataset$ can be viewed as a finite sample from its true (generally unknown) underlying distribution $\dist\coloneq p^\ttru(x,y)$, collected through an \emph{observation mechanism} governed by $\obsindic$. In particular, let $\obsindic_i \in \{0, 1\}$ indicate whether a population instance $\left(X_i, Y_i\right) \sim \dist$ is observed (and part of $\odataset$), and define the class-conditional \emph{observation propensity} 
$$\obsprb_y(x) = \prob(\obsindic = 1 \mid X = x, Y = y), \quad y \in \lbrace 0,1\rbrace.$$ 
The resulting \emph{observed distribution} of $\odataset$ is $\odist\coloneq p^\tobs(x,y) =  \prob(X=x, Y=y \mid \obsindic = 1)$ which satisfies $p^\tobs(x,y) \varpropto p^\ttru(x,y)\obsprb_y(x)$.

Class-count imbalance occurs when $p^\tobs(x,1) \ll p^\tobs(x,0)$, with $\obsprb_0(x) = \const\cdot\obsprb_1(x)$ for some constant $\const$. In other words, the probability of observing any instance is \emph{uniformly} (by a factor of $\const$) higher for class $Y=0$ than $Y=1$. The major challenge in class-count imbalance is thus the uniform shift of classification probabilities caused by observing an uneven number of instances across classes. 

In contrast, LMI occurs when $p^\tobs(x,1) \ll p^\tobs(x,0)$ and there does not exist any constant $\const: \obsprb_0(x) = \const\cdot\obsprb_1(x), \forall x \in X$. Under LMI, minority instances from particular regions of $\featspace$ are systematically less-observed than other regions, with $\obsprb_1(x) \neq \obsprb_1(x')$ for some $x,x' \in X$. The observed dataset $\odataset$ thus exhibits both a marginal imbalance $\prob(Y=1 \mid \obsindic = 1)$ and a feature-dependently distorted minority distribution $p^\tobs(x \mid y = 1)$ relative to $p^\ttru(x \mid y = 1)$. Importantly, this distortion is distinct from natural heterogeneity in $\dist$, as minority instances may be naturally rare in particular regions with small $p^\ttru(x \mid y = 1)$. However, LMI refers to an additional class-and-feature-dependent suppression through the observation mechanism $\obsprb_y(x)$ that affects minority representation in $\odataset$ despite their natural probability of occurrence $\prob(X=x \mid Y=1)$. 

Under LMI, learning from $\odataset$ faces two coupled challenges: the minority class is rare with $p^\tobs(x,1) \ll p^\tobs(x,0)$, \emph{and} the observed minority is a biased subset of its true distribution $\dist$. Therefore, correcting only class-count imbalance---the main objective of most imbalance mitigation methods---may overfit to $\odist$ and generalize poorly to $\dist$. This presents a fundamental flaw in imbalanced learning for real-life settings affected by LMI, potentially generating misleading conclusions over the relationship between $X$ and outcomes $Y$. 

\section{Proposed Methodology}

\begin{figure*}
\centering
\includegraphics[width=\textwidth,keepaspectratio]{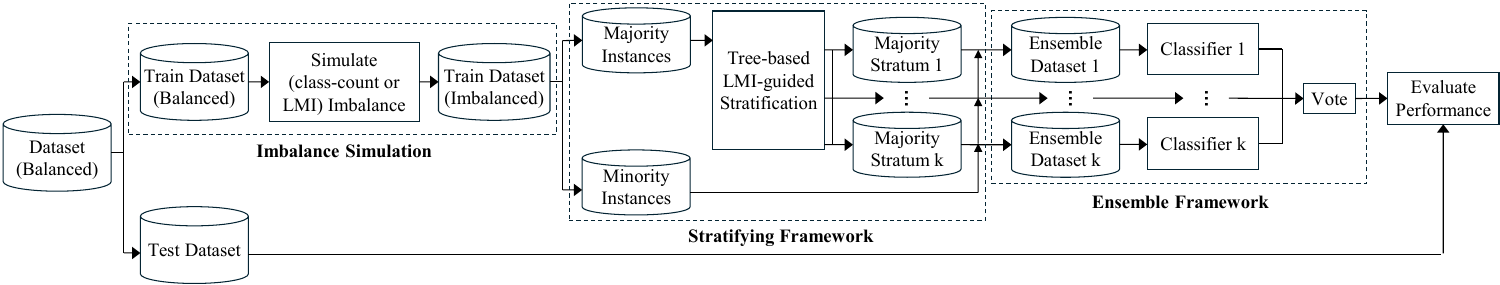}
\caption{Overall stratified ensembling (SE) framework and evaluation protocol.}
\label{fig:algo}
\end{figure*} 

We give the theoretical groundwork of our proposed Stratified Ensembling (SE) algorithm in \S\ref{sec:theory}. SE exhibits a nuanced bias-variance tradeoff which we discuss and address in \S\ref{sec:biasvariance}. Building on these theoretical results, we develop SE in \S\ref{sec:method}, and establish our proposed evaluation protocol in \S\ref{sec:protocol}. The entire proposed framework for SE is depicted in Figure~\ref{fig:algo}.

\subsection{Theoretical Motivation} \label{sec:theory}

The central challenge under LMI is not only that minority instances are scarce, but that the observed relationship between class $Y$ and features $X$ in $\odataset$ can be systematically distorted by the observation mechanism $\obsindic$. As such, models trained on $\odataset$ may learn patterns that are specific to the observed distribution $\odist$ rather than the true underlying distribution $\dist$. 

To address this, we first characterize LMI distortion in logit form, as it cleanly separates natural prevalence densities $p^\ttru(x,y)$ from observation effects $\obsindic$. Lemma~\ref{lem:obsprob} shows how class-conditional observation propensities $\obsprb_y(x)$ distort the posterior $\prob(Y=1 \mid X = x)$ into its observed counterpart $\prob(Y=1 \mid X=x, \obsindic = 1)$ on the logit (lgt) scale. 
\begin{lemma}\label{lem:obsprob}
    For any feature value $x$, the posterior class probability under the observed distribution $\odist$ satisfies $\logit\left(\prob\left(Y=1 \mid X=x, \obsindic = 1\right)\right) = \logit\left(\prob\left(Y=1\mid X=x\right)\right) + \gamma(x)$, $\gamma(x) = \log\left(\frac{\obsprb_1(x)}{\obsprb_0(x)}\right)$.
\end{lemma}
By Lemma~\ref{lem:obsprob}, the observed odds are tilted by the ratio of observation propensities
$$\frac{\prob(Y=1 \mid X=x, \obsindic = 1)}{\prob(Y=0 \mid X=x, \obsindic = 1)} = \frac{\obsprb_1(x)}{\obsprb_0(x)}\cdot\frac{\prob(Y=1 \mid X=x)}{\prob(Y=0 \mid X=x)}.$$
Moreover, the observed class-conditional feature distribution is obtained by reweighting the true underlying distribution by the observation propensity
$$p^\tobs(x\mid Y=y) \varpropto p^\ttru(x\mid Y=y)\obsprb_y(x), y \in \lbrace 0, 1 \rbrace.$$
When $\obsprb_0(x) = \const\cdot\obsprb_1(x)$---as in class-count imbalance---the selection term collapses into the constant $\const$, and the observed posterior is uniformly shifted compared to the true population posterior. As such, class-count imbalance induces a constant shift in posterior log-odds, which can be addressed, for example, by reweighting the training sample. Conversely, under LMI, the feature-dependent shift $\gamma(x)$ distorts learning non-uniformly across different regions of the feature space. 

The ideal approach to address LMI is to directly learn $\gamma(x)$ and mitigate it. However, as established in Lemma~\ref{lem:impos}, these shifts cannot be separated from true posterior $\prob(Y=1 \mid X = x)$ without using prior knowledge (which we do not assume access to).
\begin{lemma}\label{lem:impos}
Given only $\odataset$, $\prob(Y=1 \mid X = x)$ and $\gamma(x)$ cannot be separately identified.
\end{lemma}
Despite this restriction, we propose to stratify the feature space $\featspace$ into strata $\strata = \left\lbrace\strat_1, \dots, \strat_{|\strata|} \right\rbrace$ where $\gamma(x)$ is \emph{approximately} constant within any stratum $\strat_k$. As each stratum $\strat_k \subseteq \featspace$ is a relatively LMI-unbiased sample of its region and closer to class-count imbalance, it allows us to effectively address its imbalance via existing mitigation techniques. Theorem~\ref{thm:stratify} establishes the best-case scenario in which such strata exactly address LMI distortion.
\begin{theorem}\label{thm:stratify}
    Assume there exist strata $\strata$ such that for any stratum $\strat_k \in \strata$, $\gamma^k(x) = \const^k$ for some stratum-specific constant $\const^k$. Then for all $x\in\strat_k$, the Bayes-optimal ranking of instances within $\strat_k$ is identical under $\odist$ and $\dist$.
\end{theorem}
Theorem~\ref{thm:stratify} shows that it is possible to convert the complex, feature-dependent distortion caused by LMI into a simpler stratum-dependent offset that is easier to accommodate for standard learners and existing imbalance mitigation techniques. However, the assumption of Theorem~\ref{thm:stratify} is unlikely to be achieved in practice without prior knowledge on the observation mechanism $\obsindic$. In such cases, Proposition~\ref{prop:approx_strat} shows that even if $\gamma^k(x)$ is only approximately constant within a stratum $\strat^k$, the Bayes-optimal rankings are preserved for pairs of instances that are not near-ties in their true posterior log-odds.
\begin{proposition}\label{prop:approx_strat}
    Suppose that for a stratum $\strat_k \in \strata$, we have $\gamma^k(x)=\const^k+\delta^k(x)$, where $\delta^k(x)$ is the within-stratum deviation from the stratum-specific distortion constant $\const^k$. The Bayes-optimal ranking of instances within $\strat_k$ is preserved for any pair of instances whose true posterior log-odds differ by more than $2\underset{x \in \strat^k}{\max} |\delta^k(x)|$.
\end{proposition}
Proposition~\ref{prop:approx_strat} shows that the exact condition of Theorem~\ref{thm:stratify} is not necessary for stratification to be useful in mitigating LMI. Accordingly, SE does not require or assume that its learned strata have exactly constant LMI distortion. The goal is instead to reduce within-stratum variation in $\gamma(x)$ sufficiently to limit LMI-induced ranking errors. The important implication is the magnitude of the remaining within-stratum deviation $\delta^k(x)$. When deviation from the stratum-specific distortion constant $\const^k$ is small, the observed distribution can alter only the ranking of instances whose true posterior log-odds are already close. Thus, reducing the within-stratum variation of $\gamma^k(x)$ limits the set of pairs for which LMI can induce ranking errors. This provides the theoretical motivation for our stratifying procedure: the goal is not to recover perfectly unbiased strata, but to construct strata in which the residual LMI distortion is sufficiently small.

\subsection{Bias-Variance Tradeoff in Stratification}\label{sec:biasvariance}

Although stratification can reduce LMI-induced distortion, it also increases variance as each stratum contains fewer instances. This brings a nuanced bias-variance tradeoff into the procedure, which we address in two ways. 

First, instead of restricting each stratum $\strat_k$ to its few minority instances $(x_i \mid y_i = 1) \in \strat_k$, we combine all observed minority instances $(x_i \mid y_i = 1) \in \odataset$ with the majority instances $(x_i \mid y_i = 0) \in \strat_k$ of that stratum. Using all minority points in each stratum lowers variance for the minority class, while still preserving the core benefit of stratifying. In particular, the majority instances paired with minority points are drawn from the specific region of a stratum $\strat_k$ where the LMI distortion for $Y=1$ is approximately constant. Therefore, the decision boundary is less distorted by majority points of regions with a different, and potentially much more severe, LMI. 

Importantly, the primary benefit of stratifying under LMI bias is not due to minority points from other strata, and rather the disproportionate influence of majority observations in regions where minority instances are under-observed. Such majority points appear as abundant evidence for class $Y=0$ in parts of the space where $Y=1$ actually occurs but is selectively unobserved. Therefore, including such majority points systematically distorts the classifier toward predicting $\hat{y}_i=0$. By restricting the majority class to $\strat_k$, we exclude precisely these harmful majority instances from mismatched strata, while the pooled minority set mainly acts to stabilize estimation of the decision boundary. As we later show in our ablation studies in Appendix~\ref{sec:res:abl}, classifiers trained on such strata have better predictive performance compared to classifiers trained only on minority points within their associated stratum.

As our second approach for an effective bias-variance tradeoff, we ensemble the base classifier $\classifier$ trained on strata $\strata = \left\lbrace\strat_1, \dots, \strat_k \right\rbrace$, and use a performance-weighted average of probabilities 
$$\hat{\prob}(x)=\frac{\sum\limits_{\strat_k \in \strata} \alpha_k\prob_{\classifier(\strat_k)}(x)}{\sum\limits_{\strat_k \in \strata} \alpha_k}$$
over the entire training set, and not just stratum-specific instances. Here, $\alpha_k$ is an internal validation-based reliability weight, computed using a metric such as MCC, F1, or AUC. These weights are used only to reduce the influence of unstable base learners arising from differences in stratum size or local separability. They should not be, however, interpreted as unbiased estimates of target-population performance under $P^{tru}$. Values $\prob_{\classifier(\strat_k)}(x)$ are classification probabilities of the base classifier $\classifier$ trained on stratum $\strat_k$. Ensembling is well-known to reduce variance, and hence, the resulting predictions are expected to be more robust than any single classifier $\classifier(\strat_k)$.

\subsection{LMI-guided Stratified Ensembling Algorithm} \label{sec:method}

Our SE algorithm is an approximate procedure motivated by Proposition~\ref{prop:approx_strat}, that aims to create strata with reduced LMI. To this end, we develop a custom objective function that minimizes an approximate measure of within-node LMI. Using this objective function, our stratifying tree initializes with all instances in the root node and recursively partitions the feature space $\featspace$ along features $X^j \in \featspace$. The procedure terminates when further splitting a parent node is not estimated to reduce LMI distortion in its children. The resulting leaves define strata $\strata=\{\strat_1,\ldots,\strat_{|\strata|}\}$.

As our first step, we generate a one-dimensional ranking score $\prox(x)$ that orders minority probabilities according to $\prob(Y=1 | X=x, \obsindic = 1)$. The only goal of scores $\prox(x)$ is to create inexpensive and statistically stable partitions of instances $(x_i, y_i) \in \odataset$ that can be used to compare observed minority prevalence across comparable regions of the feature space. Without reducing $\featspace \in \mathbb{R}^d$ to a single ordering $\prox(x)$, partitioning the feature space becomes expensive and typically unstable in high dimensions. Given this role, the ranking scores do not require calibration, do not directly determine the feature-space partitions in SE, and are not used by the final base classifiers. In our primary implementation, we use an inexpensive closed-form difference-in-means score for $\prox(x)$, while they can also be set, for example, as the predicted sample confidence scores of any auxiliary classifier fit on $\odataset$. As we show in Appendix~\ref{sec:res:abl}, SE is robust to substantially different choices for $\prox(x)$, including an uninformative random ordering.

Next, at any node $\node$, we define $\binset_\node = \lbrace\bin_1, \dots, \bin_{|\binset_\node|}\rbrace$ as quantile bins of $\prox(x)$. For each bin $\bin$, the observed within-bin minority probability is given by 
$$\quant_\node(\bin) \coloneq \prob(Y=1 \mid X \in \featspace_\node, \prox(X) \in \bin, \obsindic = 1).$$
These probabilities can be approximated, for example, via a simple fraction of minority counts $\nummin_\bin / (\nummin_\bin + \nummaj_\bin)$ within each bin, similar to general decision tree classifiers. The heterogeneity of the observed minority prevalence at node $\node$ can then be measured via a weighted variance of $\quant_\node(\bin)$ on the logit scale:
$$\sigma_{\binset_\node}^2(\logit(\quant_\node(\bin))) = \frac{\sum\limits_{\bin \in \binset_\node} |\bin| \left(\logit\left(\quant_\node(\bin)\right) - \frac{\sum\limits_{\bin' \in \binset_\node}|\bin'|\logit\left(\quant_\node(\bin')\right)}{\sum\limits_{\bin' \in \binset_\node}|\bin'|}\right)^2}{\sum\limits_{\bin \in \binset_\node} |\bin|}.$$
Larger values of $\sigma_{\binset_\node}^2(\quant)$ indicate higher variation of observed minority prevalence across ranking regions. However, as previously discussed, this can be due to feature-dependent LMI distortion caused by $\gamma(x)$, or natural heterogeneity of the minority distribution $p^\ttru(x\mid y = 1)$ at that region. While it is not possible to exactly separate these components without prior knowledge on the observation mechanism $\obsindic$, we propose a compromise that takes advantage of the additive log-odds structure established in Lemma~\ref{lem:obsprob}.  

Instead of directly using values $\sigma_{\binset_\node}^2(\quant)$ to measure LMI, we use a parent-child relationship of our stratifying tree to approximately isolate LMI distortion from natural minority prevalence. In particular, we first enforce the child and parent nodes of our stratifying tree to use the same binning partitions. That is, after splitting a parent node $\node$ into left and right child nodes $\child_l, \child_r$, each child inherits the parent bins corresponding to its specific region. This shared-binning rule ensures that any parent-child comparison is performed over similar neighborhoods, rather than on differences induced by changing bin definitions.

For any parent node $\node$, child node $\child$, and shared bin $\bin \in \binset_\node \cap \binset_\child = \binset_\child$, define the residual
$$
\Delta_{\node \to \child}(\bin) = \logit\left(\quant_\node(\bin)\right) - \logit\left(\quant_\child(\bin)\right).
$$
Residuals $\Delta_{\node \to \child}(\bin)$ can be theoretically decomposed into a parent-child drift of natural prevalence and LMI distortion as given in Theorem~\ref{thm:split}.
\begin{theorem}\label{thm:split}
    Residuals $\Delta_{\node \to \child}(\bin)$ are composed of a sum of natural-prevalence drift \newline$\logit\left(P(Y=1\mid x\in \featspace_\node,\eta(x)\in \bin)\right) - \logit\left(P(Y=1\mid x\in \featspace_\child,\eta(x)\in \bin)\right)$ and LMI-distortion drift $\gamma_\node(\bin)-\gamma_\child(\bin)$.
\end{theorem}
Given this decomposition, Corollary~\ref{cor:decomp} shows that when parent and child nodes have similar natural heterogeneity on the logit scale and within a bin $\bin \in \binset_\child$, residuals $\Delta_{\node \to \child}(\bin)$ primarily reflect differences in LMI-distortion.
\begin{corollary}\label{cor:decomp}
    If the within-bin probabilities $\quant_\node(\bin)$ of a bin $\bin \in \binset_\child$ shared among parent $\node$ and child $\child$ do not substantially change the logit of the true posterior $\logit\left(P(Y=1\mid x\in \featspace_\node,\eta(x)\in \bin)\right)$ then residuals $\Delta_{m\to v}(b)$ primarily reflect LMI-distortion drift.
\end{corollary}
As $\gamma(x)$ is not identifiable from $\odataset$, this decomposition does not allow us to certify that a learned stratum has approximately constant LMI distortion. Rather, Corollary \ref{cor:decomp} provides a conditional link between the observable residual and the target distortion. When natural-prevalence drift within shared bins is small, variation in the residual primarily reflects variation in LMI distortion. Our SE algorithm thus uses this residual as an observed-data surrogate rather than as an exact estimator of \(\gamma(x)\).

Specifically, we use $\Delta_{\node\to\child}(\bin)$ to define a node-level objective function that favors partitions with more homogeneous parent-child prevalence changes across shared bins, which may help identify regions with reduced LMI distortion under the conditions discussed above. At a node $\node$, a candidate split is specified by a feature $X^j$ and a threshold among the ones defined by bin set $\binset_\node$. For the candidate split (feature and threshold) we compute the weighted score
$$ \frac{|\child_l|}{|\node|}\,\sigma_{\binset_{\child_l}}^2(\Delta_{\node\to\child_l})+\frac{|\child_r|}{|\node|}\,\sigma_{\binset_{\child_r}}^2(\Delta_{\node\to\child_r}),
$$
and choose the split that minimizes it (equivalently maximizing the gain from the parent), subject to a stopping criterion based on the parent node's observed heterogeneity. The recursion terminates when no feasible split yields a gain. 

In the case that Corollary~\ref{cor:decomp} is not entirely satisfied, the stratifying algorithm may select a feature that reduces natural-prevalence drift more than LMI-distortion drift. Given our bias-variance tradeoff approach in \S\ref{sec:biasvariance}, such a split corresponds to the general undersampling of the majority class in the resulting children. While such splits may not reduce LMI-distortion, they also do not deteriorate the condition of Proposition~\ref{prop:approx_strat}, as shown in Proposition~\ref{prop:deter}.
\begin{proposition}\label{prop:deter}
     For any pair $x,x'\in \child$, the ranking-stability condition in Proposition~\ref{prop:approx_strat} is weakly no harder to satisfy in the child $\child$ than in the parent $\node$.
\end{proposition}
In practice, over-partitioning is further constrained as a split is accepted only when it improves the SE objective and satisfies the minimum leaf-size requirements. Moreover, each resulting classifier reuses the full observed minority set while ensembling reduces the variance introduced by partitioning, as discussed in \S\ref{sec:biasvariance}. Appendix \ref{sec:res:abl} further shows that performance remains stable when the minimum leaf sizes and score-bin count are varied.

\subsection{Gold-Standard Evaluation Protocol} \label{sec:protocol}

An important challenge in LMI class imbalance is that sampling test sets from datasets that are potentially LMI-imbalanced biases true model performance. To highlight and address this potential bias, we propose to measure imbalance performance using gold-standard datasets that are originally near-balanced and unbiased, and simulate LMI or class-count imbalance on those datasets. That is, we first partition the dataset into training $\traindata$ and test $\testdata$ sets using stratified folds. We then simulate imbalance via controlled instance deletions---based on features $X$ and class $Y=1$ for LMI, and randomly for class-count imbalance. The simulation algorithms are given in Appendix~\ref{app:algos:deletion}. 

All methods are applied to the resulting imbalanced training set, while performance is evaluated on the gold-standard test set. This design ensures that comparisons across methods reflect their ability to recover the predictive power over the true underlying distribution $\dist$, rather than the biased $\odist$.

Our simulation-based evaluation is necessary as true-LMI distorted datasets are difficult to obtain, particularly as benchmarks for imbalance learning. Therefore, we emphasize that our protocol is not meant to replace current evaluation practices. Rather, we seek to highlight the risks of benchmarking methods solely on test sets drawn from potentially LMI datasets. In practice, gold-standard test sets can be specifically generated by controlling the data collection procedure, for example, via targeted labeling, external cohorts, delayed-outcome follow-ups, or auditing data collection. 

\section{Numerical Evaluation} \label{sec:res}

Our experiments evaluate (1) the performance of our approach versus benchmarks under LMI; and (2) how standard hold-out evaluation can yield misleading results when applied to LMI data. We provide further numerical results in Appendix~\ref{app:res}, including time and memory usage, varying degrees of LMI, class-specific and additional evaluation metrics, detailed non-averaged results, ablation and hyperparameter sensitivity studies, alternative LMI mechanisms, and performance under general class-count imbalance.

\subsection{Evaluation Framework} \label{sec:res:eval}

We construct 5-fold cross-validation hold-out samples taken from benchmark datasets of the KEEL repository \citep{alcala2011keel_dataset}, detailed in Appendix~\ref{sec:res:data}. Then, on training data $\traindata$, we simulate $95\%$ deletion or equivalently a 1:20 IR (corresponding to mildly severe imbalance) with either class-count or LMI simulated according to the procedure detailed in Appendix~\ref{app:algos:deletion}. Following prior work \citep{rezvani2023broad,altalhan2025imbalanced}, we report MCC, and F1 as our main criteria, but also report other metrics in Appendix~\ref{sec:res:othermet}.  

We evaluate on five base classifiers and a host of popular and state-of-the-art imbalance methods detailed in Appendix~\ref{app:algos}. All algorithms and classifiers are tuned via a grid search and 5-fold cross-validation performed on the training set. We also report two reference empirical baselines. The gold standard baseline (GLD): performance on the original balanced datasets, and the damaged baseline (DMG): performance on imbalanced datasets without any imbalance correction technique. These baselines contextualize the gains achieved by different methods. 

All algorithms are coded in Python, occasionally with a C++ backend (all code is provided in the supplementary materials) and executed on a PC with an Intel Xeon processor W-2255, 264GB of RAM, and Ubuntu 20.04.5 operating system. The SE code is \href{https://github.com/aminhn/SE}{available} and open source.

\subsection{Performance under LMI} \label{sec:res:mnar}

\begin{figure*}
     \centering
     \subfloat[][\centering Overall performance on unbiased hold-out sets]{
         \centering
         \includegraphics[scale=0.3]{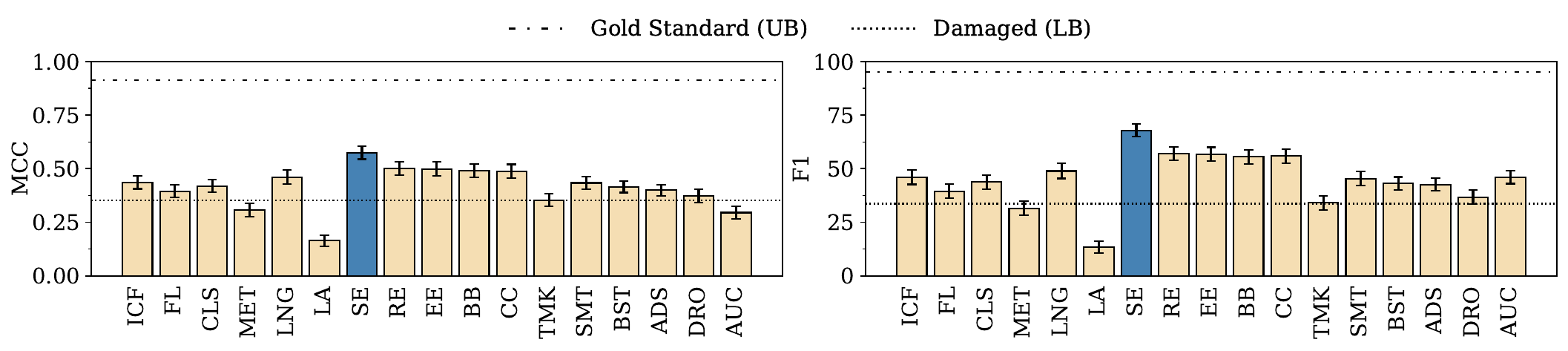}
         \label{fig:main_ovr}
     }
     
     \subfloat[][\centering Performance on low-LMI regions]{
         \centering
         \includegraphics[scale=0.3]{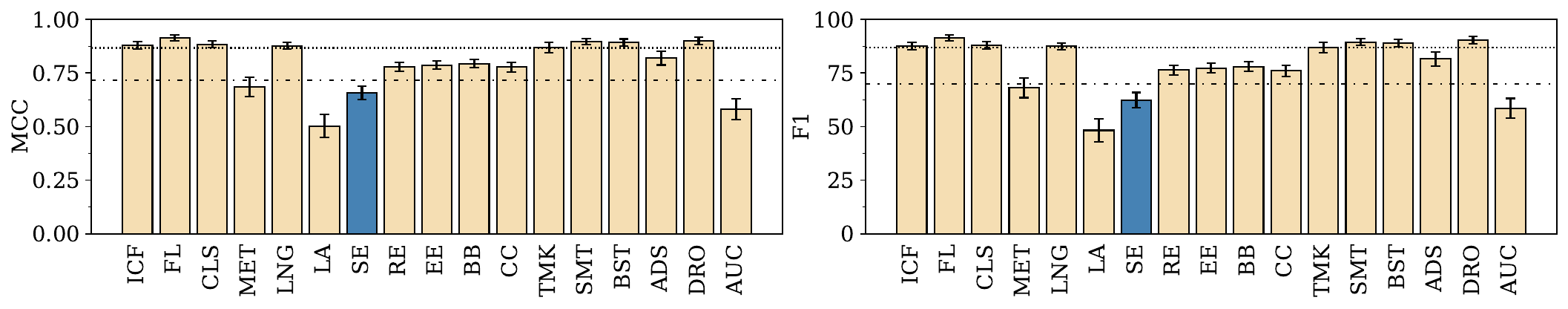}
         \label{fig:main_rem}
     }
     
    \subfloat[][\centering Performance on high-LMI regions]{
         \centering
         \includegraphics[scale=0.3]{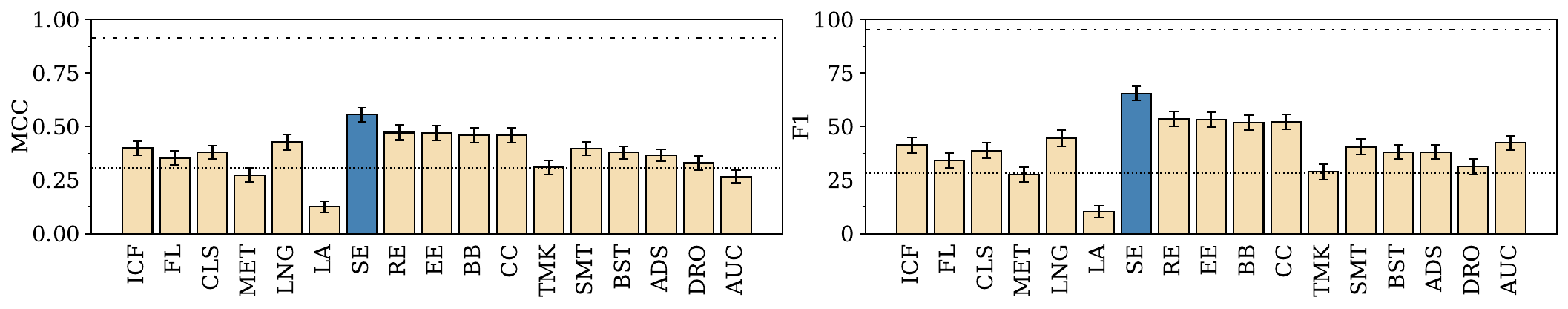}
         \label{fig:main_del}
     }
    \caption{Average performance under LMI across datasets and base classifiers.}
    \label{fig:main}
\end{figure*}

Figure~\ref{fig:main} gives the MCC and F1 under LMI. All classifiers show notable performance drops when trained on damaged data (DMG lower bound in Figure~\ref{fig:main_ovr}), confirming the adverse effect of LMI. SE achieves the strongest overall performance on the undamaged test population, driven particularly by improved generalization to high-LMI regions (Figure~\ref{fig:main_del}). On the other hand, other methods overfit the majority class in under-observed regions of the feature space, giving misleadingly high performance in low-LMI regions (Figure~\ref{fig:main_rem}) but considerably low performance on high-LMI regions. 

The misleading advantage of competing methods is itself evidence of overfitting to the biased LMI observation pattern. In particular, DMG---trained directly on the damaged data without any imbalance correction---substantially exceeds GLD in these regions, even though GLD is trained on the original unbiased data. At the same time, DMG performs considerably worse on both the full population and in high-LMI regions. Therefore, its higher score in Figure~\ref{fig:main_rem} does not reflect a superior target-population decision rule. Rather, it reflects overfitting to the regions overrepresented by the LMI observation mechanism. By contrast, SE reduces overfitting and remains close to GLD in low-LMI regions, substantially improves performance in high-LMI, and achieves the highest overall MCC and F1 on the full undamaged test data.

Lastly, SE also substantially outperforms the other ensembling baseline methods (RE, EE, BB), suggesting that its feature-guided stratification strategy provides benefits beyond those obtained by conventional random undersampling and ensembling.

\subsection{Biased Evaluation by Sampling LMI Data}

\begin{figure*}
\centering
\includegraphics[scale=0.3]{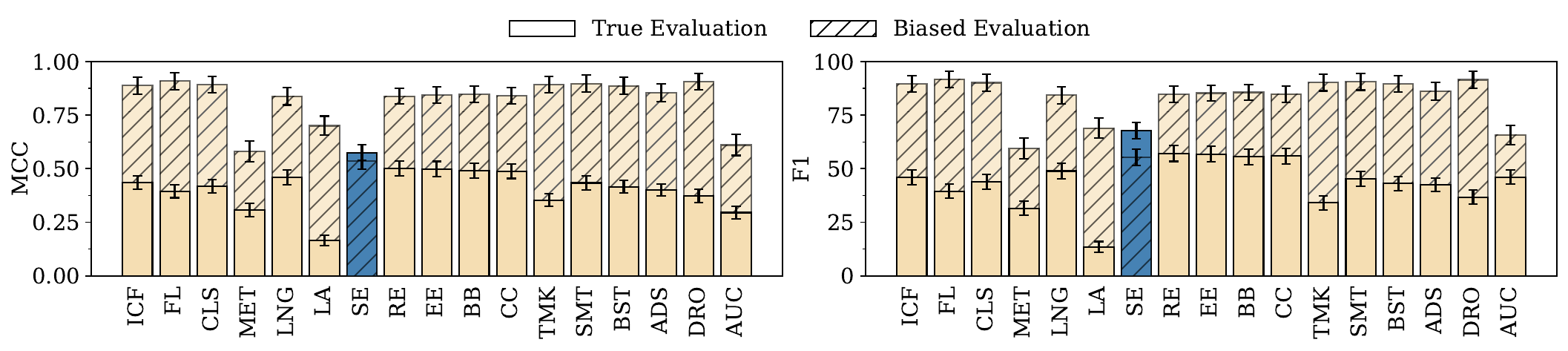}
\caption{True versus biased evaluation by conventional sampling of LMI-biased hold-out sets}
\label{fig:dmgeval}
\end{figure*} 

Standard evaluation protocols will mislead model performance if they sample the test sets from the same LMI-effected data. To reveal this flaw, we simulate $85\%$ LMI (to retain sufficient instances for subsequent hold-out samples), and evaluate models on hold-out sets drawn solely from the damaged dataset (mimicking conventional practice) to end at $95\%$ LMI. This setup mirrors real-world conditions where the test data shares the same structural bias as training, but it fails to reflect the true minority population under LMI. 

Figure~\ref{fig:dmgeval} reports performance average results. Under LMI, evaluating on data drawn from the same biased distribution used for training can overstate performance and obscure generalization. In particular, this biased evaluation misleadingly favors models that overfit to the observed minority subset while penalizing those that generalize to the entire population. These results reinforce the importance of accounting for the nature of imbalance when evaluating mitigation methods.

\section{Conclusion} \label{sec:conclusion}

This paper studies LMI class imbalance, where minority observability depends jointly on class labels and predictive features. Under LMI, popular imbalance methods fail to generalize and instead amplify biases in the observed data. To address this, we develop a stratifying procedure that uses an observed-data surrogate to seek subsets with reduced within-stratum LMI distortion. These strata are then used to create an ensemble of classifiers that are trained on the entire set of observed minority instances plus the majority instances of a generated stratum. Our method outperforms popular and state-of-the-art techniques under LMI on multiple criteria. We also introduce an evaluation protocol that reveals how standard validation practices can misrepresent performance under LMI.

\bibliography{iclr_bib}
\bibliographystyle{plainnat2}

\appendix
\onecolumn

\section{Proofs}

\subsection*{Proof of Lemma~\ref{lem:obsprob}}

By Bayes' rule, for $y\in\{0,1\}$ we have
\[
\prob(Y=y \mid X=x, O=1)
=\frac{\prob(O=1\mid X=x, Y=y)\prob(Y=y\mid X=x)}{\prob(O=1\mid X=x)}.
\]
Taking the odds ratio under the observed distribution yields
$$
\frac{\prob(Y=1 \mid X=x, O=1)}{\prob(Y=0 \mid X=x, O=1)} =
\frac{\prob(O=1\mid X=x, Y=1)\prob(Y=1\mid X=x)}{\prob(O=1\mid X=x, Y=0)\prob(Y=0\mid X=x)} 
$$
$$
= \frac{\obsprb_1(x)}{\obsprb_0(x)}\cdot
\frac{\prob(Y=1\mid X=x)}{\prob(Y=0\mid X=x)}.
$$
Applying $\log$ to both sides gives
\[
\log\left(\frac{\prob(Y=1 \mid X=x, O=1)}{\prob(Y=0 \mid X=x, O=1)}\right)
=
\log\left(\frac{\prob(Y=1\mid X=x)}{\prob(Y=0\mid X=x)}\right)
+\log\left(\frac{\obsprb_1(x)}{\obsprb_0(x)}\right).
\]
Given $\logit(p) = \log\left(\frac{p}{1-p}\right)$, and by defining $\gamma(x)\coloneq \log\left(\frac{\obsprb_1(x)}{\obsprb_0(x)}\right)$ proves the claim.

\subsection*{Proof of Lemma~\ref{lem:impos}}
For any function $h(x)$, we can write $\ell^\tobs(x)=\ell(x)+\gamma(x)=(\ell(x)+h(x))+(\gamma(x)-h(x))$. Thus the same observed posterior log-odds are consistent with infinitely many decompositions into a natural log-odds component and an observation-distortion component. Therefore, $\gamma(x)$ cannot be recovered from $\ell^\tobs(x)$ alone without additional structure.

\subsection*{Proof of Theorem~\ref{thm:stratify}}

Fix any stratum $\strat_k\in\strata$, and let 
\[
\beta^{\ttru}(x)\coloneq\prob(Y=1\mid X=x), \quad
\beta^{\tobs}(x)\coloneq\prob(Y=1\mid X=x,\obsindic=1),
\]
denote the true and observed posteriors, respectively. By Lemma~\ref{lem:obsprob}, for any $x \in \strat_k$ we have,
\[
\logit\big(\beta^{\tobs}(x)\big)=\logit\big(\beta^{\ttru}(x)\big)+\gamma(x),
\qquad
\gamma(x)=\log\left(\frac{\obsprb_1(x)}{\obsprb_0(x)}\right).
\]
By the statement, $\forall x \in \strat_k$ we have $\gamma(x) = \const^k$, and hence
$$
\beta^{\tobs}(x) = \logit^{-1}\left(\logit\big(\beta^{\ttru}(x)\big)+\const^k\right),
$$
where $\logit^{-1}(t)=1/(1+e^{-t})$ is the inverse-logit function. 

Since both $\logit(\cdot)$ and $\logit^{-1}(\cdot)$ are strictly increasing, the mapping
\[
f_k(u)\coloneq\logit^{-1}\left(\logit(u)+\const^k\right),\qquad u\in(0,1),
\]
is strictly increasing in $u$. Therefore, for any two points $x,x'\in\strat_k$ that
\[
\beta^{\ttru}(x)>\beta^{\ttru}(x')
\Longrightarrow
f_k\big(\beta^{\ttru}(x)\big)>f_k\big(\beta^{\ttru}(x')\big)
\Longrightarrow
\beta^{\tobs}(x)> \beta^{\tobs}(x'),
\]
Since the Bayes-optimal ranking under either distribution is obtained by sorting instances by their posterior probabilities, the Bayes-optimal ranking within $\strat_k$ is identical under $\odist$ and $\dist$.

\subsection*{Proof of Proposition~\ref{prop:approx_strat}}

Fix any stratum $\strat_k\in\strata$, and let 
\[
\beta^{\ttru}(x)\coloneq\prob(Y=1\mid X=x), \quad
\beta^{\tobs}(x)\coloneq\prob(Y=1\mid X=x,\obsindic=1),
\]

By Lemma~\ref{lem:obsprob} and the statement we have
\[
\logit\left(\beta^{\tobs}(x)\right) = \logit\left(\beta^{\ttru}(x)\right) + \const^k +\delta^k(x).
\]
For any $x,x'\in\strat_k$ we thus have 
$$\logit\left(\beta^{\tobs}(x)\right) - \logit\left(\beta^{\tobs}(x')\right) = \logit\left(\beta^{\ttru}(x)\right) - \logit\left(\beta^{\ttru}(x')\right) + \delta^k(x)-\delta^k(x').$$

Since $|\delta^k(x)|\leq \underset{x \in \strat^k}{\max} |\delta^k(x)|$ and $|\delta^k(x')|\leq \underset{x' \in \strat^k}{\max} |\delta^k(x')|$, we have $|\delta^k(x) -\delta^k(x')|\leq 2\underset{x \in \strat^k}{\max} |\delta^k(x)|$.
Therefore, if
\[
\left|
\logit\left(\beta^{\ttru}(x)\right) - \logit\left(\beta^{\ttru}(x')\right) \right|
>
2\underset{x \in \strat^k}{\max} |\delta^k(x)|,
\]
the residual term cannot change the sign of the true log-odds difference, and hence, the observed and true rankings agree.

\subsection*{Proof of Theorem~\ref{thm:split}}

By definition, we have
\[
\quant_\node(\bin)=\prob(Y=1\mid X\in \featspace_\node,\ \prox(X)\in \bin,\ \obsindic=1),
\]
\[
\quant_\child(\bin)=\prob(Y=1\mid X\in \featspace_\child,\ \prox(X)\in \bin,\ \obsindic=1).
\]

By applying Lemma~\ref{lem:obsprob}, 
$$
\logit\big(\quant_\node(\bin)\big) =
\logit\big(\prob(Y=1\mid X\in \featspace_\node,\ \prox(X)\in \bin)\big)+\gamma^{\node}(\bin),\label{eq:parent}
$$
$$
\logit\big(\quant_\child(\bin)\big) =
\logit\big(\prob(Y=1\mid X\in \featspace_\child,\ \prox(X)\in \bin)\big)+\gamma^{\child}(\bin).\label{eq:child}
$$
Subtracting the terms gives
\begin{align*}
\Delta_{\node \to \child}(\bin) &=
\Big[
\logit\big(\prob(Y=1\mid X\in \featspace_\node,\ \prox(X)\in \bin)\big)
-
\logit\big(\prob(Y=1\mid X\in \featspace_\child,\ \prox(X)\in \bin)\big)
\Big]
\\&+ \big[\gamma^{\node}(\bin)-\gamma^{\child}(\bin)\big].
\end{align*}
The first term corresponds to natural-prevalence drift, and the second term to LMI-distortion drift.

\subsection*{Proof of Corollary~\ref{cor:decomp}}

By the assumption, for each shared bin $\bin$ the true log-odds are approximately invariant when conditioning from $\featspace_\node$ to $\featspace_\child$, i.e.,
\[
\logit\big(\prob(Y=1\mid X\in \featspace_\node,\ \prox(X)\in \bin)\big)\approx
\logit\big(\prob(Y=1\mid X\in \featspace_\child,\ \prox(X)\in \bin)\big).
\]
Therefore, the natural-prevalence drift of $\Delta_{\node \to \child}(\bin)$ is approximately zero, and we can conclude
\[
\Delta_{\node \to \child}(\bin)=\logit\big(\quant_\node(\bin)\big)-\logit\big(\quant_\child(\bin)\big)
\approx
\gamma^{\node}(\bin)-\gamma^{\child}(\bin).
\] 
In the special case where natural-prevalence drift cancels exactly, the approximation becomes equality.

\subsection*{Proof of Proposition~\ref{prop:deter}}

Let \(c^\node = \arg\underset{c}{\min} \underset{x\in \node}{\sup}|\gamma(x)-c|\) and \(c^\child = \arg\underset{c}{\min} \underset{x\in \child}{\sup}|\gamma(x)-c|\). Since
\(\child \subset \node \), we have
\[
\sup_{x\in \child}|\gamma(x)-c^\child| \leq \sup_{x\in \child}|\gamma(x)-c^\node| \leq \sup_{x\in \node}|\gamma(x)-c^\node|. \]
Thus the best achievable within-child deviation from a constant distortion is
no larger than the best achievable within-parent deviation.

\section{Evaluation Details}

We detail our benchmark datasets in Appendix~\ref{sec:res:data}, introduce baseline imbalance mitigation methods and their hyperparameters in Appendix~\ref{app:algos}, and give our overall class-count and LMI simulation procedure in Appendix~\ref{app:algos:deletion}. The pseudocode of our proposed SE is given in Appendix~\ref{app:algos:SERBE}.

\subsection{Datasets} \label{sec:res:data}

We considered all balanced benchmark datasets from the KEEL repository \citep{alcala2011keel_dataset} with more than 2,000 instances. This was to give us enough instances to simulate class-count or LMI. The datasets and their properties are given in Table~\ref{tab:data}.

\begin{table}
\centering
\caption{Datasets, their number of instances, and dimension of $X$.}
\label{tab:data}
\begin{tabular}{lcc c lcc}
\toprule
Dataset & $\numinstance$ & $\numdim$ && Dataset & $\numinstance$ & $\numdim$ \\
\midrule
page-blocks0 & 7860 & 10 && segment0 & 3166 & 19  \\
abalone19 & 6626 & 8  && shuttle-c0-vs-c4 & 2728 & 9  \\
yeast3 & 2114 & 8 && yeast4 & 2294 & 8  \\
yeast5 & 2304 & 8 && yeast6 & 2318 & 8  \\
\bottomrule
\end{tabular}
\end{table}

\subsection{Algorithms} \label{app:algos}

We evaluate on five base classifiers: logistic regression, support vector machines, random forests, multi-layer perceptrons, and CatBoost, covering linear, kernel-based, ensemble, neural, and boosting models with linear and non-linear decision boundaries. 

We consider popular and state-of-the-art imbalance methods in six categories detailed below. For fairness across algorithms, we use standard implementations (scikit-learn, imbalanced-learn, or PyTorch) and widely adopted default hyperparameters and tune more complex algorithms where stated. The implementation code of all algorithms is provided in the supplementary materials. 

\subsubsection{Cost-based Methods}

Inverse Class Frequency (ICF): Each class receives a weight inversely proportional to its training-set frequency. In particular, the minority class weight is set to $\frac{\numinstance}{2\nummin}$ and the majority to $\frac{\numinstance}{2\nummaj}$.

Focal Loss (FL): Extends cross-entropy by down-weighting easy examples \citep{lin2017focal}. The loss for a true class with predicted probability $\prob_t$ is $-\alpha_t(1-\prob_t)^\gamma \log(\prob_t)$. We use the canonical defaults $\alpha=0.25$ (minority class weight) and $\gamma=2$ (focusing parameter), and tune them via 5-fold cross-validation.

Meta Reweighting (MET) : Iteratively learns per-training-point weights so that training instances whose gradients align with (i.e., most reduce) the meta loss get higher weight while harmful updates get down-weighted. We tuned base parameters using 5-fold cross-validation.

Long-tail rebalancing with tuned priors (LNG): Trains a base classifier, and then post-hoc adjusts its predicted probability of the minority group by subtracting a log-odds shift equal to $\tau\log\left(\frac{p}{1-p}\right)$, where $p$ is the observed minority rate, to compensate for long-tail class priors. We tuned base parameters using 5-fold cross-validation.

Class-Balanced Loss (CLS): Weights classes using their ``effective number of samples'' $E(n_y)=(1-\beta^{n_y})/(1-\beta)$ \citep{cui2019class}. Class weight is proportional to $\frac{1}{E(n_y)}$. Following \citet{cui2019class}, we set $\beta=\frac{(n-1)}{n}$, the standard default and tune it using 5-fold cross-validation.

\subsubsection{Debiasing Methods}

Logit Adjustment (LA): Corrects for skewed class priors by adding $\tau \log \pi_y$ to each class logit before cross-entropy, where $\pi_y$ is the empirical class prior \citep{menon2020long}. We use the standard setting $\tau=1$ and tune it using 5-fold cross-validation.

\subsubsection{Ensemble-based Methods}

Easy Ensemble (EE): Constructs multiple balanced subsets obtained through majority undersampling of the dataset and trains AdaBoost learners on each set \citep{liu2008exploratory}. We use the \texttt{EasyEnsembleClassifier} Python library from imbalanced-learn with default hyperparameters.

Balanced Bagging (BB): Implements bagging with balanced bootstrap samples. We use the \newline \texttt{BalancedBaggingClassifier} Python library from imbalanced-learn \citep{lemaavztre2017imbalanced} with default hyperparameters.

Random Ensembling (RE): Constructs a stratified ensemble using random subsets of the dataset. Instead of using our proposed objective function to guide stratification and the stopping criteria, RE randomly undersamples the majority class until each stratum is class-count balanced. 

\subsubsection{Undersampling Methods}

Cluster Centroids (CC): Replaces the majority class with its $k$ cluster centroids, reducing its size to match the minority count \citep{yen2009cluster}. We use the  \texttt{ClusterCentroids} Python library from imbalanced-learn \citep{lemaavztre2017imbalanced} with default hyperparameters. 

Tomek Links (TMK): Undersamples the majority set by removing borderline instances that form Tomek pairs (nearest neighbors of opposite class). We use the \texttt{TomekLinks} Python library from imbalanced-learn \citep{lemaavztre2017imbalanced} with default hyperparameters. 

\subsubsection{Oversampling Methods}

SMOTE (SMT): Generates synthetic minority samples by interpolating between each minority point and its nearest neighbors \citep{chawla2002smote}. We use the \texttt{SMOTE} Python library from imbalanced-learn \citep{lemaavztre2017imbalanced} with default hyperparameters.

Borderline SMOTE (BST): A variant of SMOTE that oversamples only minority instances near the decision boundary (with many majority neighbors) \citep{han2005borderline}. We use the \texttt{BorderlineSMOTE} Python library from imbalanced-learn \citep{lemaavztre2017imbalanced} with default hyperparameters.

Adaptive Synthetic oversampling (ADS): Generates more synthetic points in regions where the minority is sparsely represented relative to the majority \citep{he2008adasyn}. We use the \texttt{ADASYN} Python library from imbalanced-learn \citep{lemaavztre2017imbalanced} with default hyperparameters.

\subsubsection{Optimization-based Methods}

AUC-PR directive objectives (AUC): iteratively refits the classifier while updating sample weights to directly target AUC-PR: for a set of recall anchors, it finds the score threshold on positives that achieves each recall, then up-weights (i) positives that fall below that threshold (misses) and (ii) negatives that exceed it (false positives), with a larger penalty on false positives when recall is high. The weights are clipped/floored, normalized, and EMA-blended each iteration for stability. We tuned base parameters using 5-fold cross-validation.

Distributionally Robust Optimization (DRO): alternates between fitting the classifier with sample weights $w=q.n$ and updating the weight distribution $q$ to put more mass on high-loss examples (via an exponential/multiplicative update), then projects $q$ back onto a capped simplex so no single point can dominate (cap $=\frac{1}{\alpha n}$). This implements a DRO worst-case reweighting over an $\alpha$-sized effective subset, with optional smoothing back toward uniform to keep all weights positive. We tuned base parameters using 5-fold cross-validation.

\subsection{LMI and Class-count Imbalance Simulation} \label{app:algos:deletion}

Algorithm~\ref{alg:mnar-damage} gives the pseudocode of our damaging procedure to simulate LMI or class-count imbalance. The damaging procedure simulates LMI via a soft deletion mechanism that incrementally removes minority instances based on a specified ordering $O^{\text{dmg}}$ of minority instances in the training data. This ordering is an intentional design, such that removing instances with lower order simulates a non-random deletion mechanism for LMI. Without such ordering, deletion becomes approximately random and closer to general class-count imbalance.

We consider six different LMI mechanisms, spanning monotonic vs non-monotonic, single vs multiple feature driven, observed vs unobserved feature driven, minority vs majority based, and linear vs categorical deletion.
\begin{itemize}[leftmargin=*]
    \item Monotonic LMI: The default deletion mechanism that orders instances by a randomly selected feature $X\in \featspace$. Instances with lower values of $X$ receive the highest deletion priority.
    \item Non-monotonic LMI: Instances are ordered by the absolute value of their percentile rank difference to the median $|(\text{rank}(X) - 0.5)|$ of a randomly selected feature $X\in \featspace$. Here, $\text{rank}(X)$ determines the rank of an instance by ordered values of $X$ in $\traindata$. Therefore, instances with higher or lower values of $X$ (tails) receive the highest deletion priority.
    \item Interaction LMI: Instances are ordered by the geometric mean of the percentile ranks $\sqrt{\text{rank}(X) \times \text{rank}(X')}$ of two randomly selected features $X, X' \in \featspace$. Instances with the lower values of this nonlinear combination of percentile ranks receive the highest deletion priority.
    \item Unobserved LMI: Instances are ordered by values of an unobserved feature. Feature $X$ is randomly selected from $\featspace$ and used only for ordering. The training and test datasets are both updated to $\featspace \setminus X$, and therefore, all mitigation procedures and base classifiers never observe $X$. Instances with lower values of $X$ receive the highest deletion priority.
    \item Majority-bias LMI: Instances are ordered by their average distance to the nearest majority points. Instances nearest to majority points have higher deletion probability.
    \item Categorical LMI: Instances are ordered by their quantile bin assignment of a randomly selected feature $X\in \featspace$. Instances in lower quintiles receive the highest deletion priority.
\end{itemize}

Feature interactions LMI makes observation propensity depend on a nonlinear interaction between two observed features, testing whether SE can recover localized structure that is not attributable to any single covariate. Unobserved-driver LMI introduces feature-dependent selection using a variable that is omitted from the observed feature set. Majority-bias LMI determines minority deletion propensity using local majority concentration, producing imbalance that depends on the surrounding class distribution rather than directly on a single minority feature. Non-monotonic LMI deletes minority instances from both tails of a driving feature, creating a distortion pattern that cannot be represented by a one-directional monotonic mechanism. Lastly, categorical LMI induces under-observation according to discretized quantile groups, testing whether SE remains effective when the observation mechanism operates over discrete rather than continuous feature regions.

When applicable, each deletion mechanism is simulated five times using a distinct, randomly selected feature(s) $X\in \featspace$, and repeated five times based on a random seed that determines the soft deletion likelihood. 

Deletion likelihoods vary smoothly with the ordering $O^{\text{dmg}}$ to ensure the possibility that some minority points remain observable across all regions. This preserves partial support of the original distribution $\dist$ while inducing systematic distortions in higher deletion-probability regions, modeling settings where imbalance arises from bias that skews but possibly does not fully censor subpopulations. Furthermore, by applying deletions progressively at increasing thresholds (65\%, 75\%, 85\%, 95\%), we generate a controlled family of datasets that allow us to measure the effects of varying degrees of LMI in Appendix~\ref{sec:res:vary}.

\begin{algorithm}
\caption{Imbalance Simulation Algorithm}
\label{alg:mnar-damage}
\begin{algorithmic}[1]
\REQUIRE Minority instances $\minset=\{i \in \traindata:y_i=1\}$, damage level $\theta^\text{tar}$, damage type $\texttt{how} \in \{\text{Class-count}, \text{LMI}\}$, ordering $O^{\text{dmg}}$.
\STATE Initialize set of damaged indices $\mathcal{I}^{\text{dam}} \gets \emptyset$
\FOR{$\theta \in \{0.65, 0.75, 0.85, 0.95\}$}
    \STATE Set $n^{\text{tar}} \gets \theta \times \left| \minset\right|$ 
    \IF{$\texttt{how} = \texttt{class-count}$} 
        \STATE Sample $n^{\text{tar}} - \left| \mathcal{I}^{\text{dam}} \right|$ points at random: $\mathcal{I}^{\text{dam}} \gets \mathcal{I}^{\text{dam}} \cup 
        \text{Sample}\left(\{i \in \minset \}, \, n^{\text{tar}}\right)$
    \ELSE 
        \STATE $\lambda \gets 0.9 - \dfrac{0.1\times \left| \mathcal{I}^{\text{dam}} \right|}{n^{\text{tar}}}$
        \FOR{Instances $i \in \minset$ according to ordering $O^{\text{dmg}}$}
            \IF{$i \notin \mathcal{I}^{\text{dam}}$ \textbf{and} $\text{Uniform}(0,1) \leq \lambda$}
                \STATE $\mathcal{I}^{\text{dam}} \gets \mathcal{I}^{\text{dam}} \cup \{i\}$
                \STATE $\lambda \gets \lambda - 0.1/n^{\text{tar}}$
            \ENDIF
            \IF{$\left| \mathcal{I}^{\text{dam}} \right| = n^{\text{tar}}$}
                 \STATE \textbf{break}
            \ENDIF
        \ENDFOR
    \ENDIF
    \IF{$\theta = \theta^\text{tar}$}
        \STATE \textbf{break}
    \ENDIF
\ENDFOR
\RETURN $\traindata \setminus \mathcal{I}^{\text{dam}}$
\end{algorithmic}
\end{algorithm}

\subsection{Stratifying Ensembling (SE) Pseudocode} \label{app:algos:SERBE}

Algorithm~\ref{alg:se} gives the pseudocode of our proposed approach.

\begin{algorithm}
\caption{Stratified Ensembling (SE) Algorithm}
\label{alg:se}
\begin{algorithmic}[1]
\REQUIRE Training set $\traindata=\{(x_i,y_i)\}_{i=1}^n$, $\majset=\{i:y_i=0\}$, $\minset=\{i:y_i=1\}$; base learner $\classifier$; Number of root bins $|\binset_0|$; min leaf size $n_{\min}$; Validation metric $\mathrm{Perf}(\cdot)$
\STATE Compute proxy scores $\prox(x_i)$ for all $i$ 
\STATE Form global $|\binset_\node|$ quantile bins $\binset=\{\bin_1,\dots,\bin_{|\binset_0|}\}$
\STATE Grow a stratifying tree on $X$ to obtain leaf strata $\strata=\{\strat_1,\dots,\strat_K\}$ as follows:
\FOR{node $\node$ (top-down), $|\node|\ge n_{\min}$}
    \STATE For each $\bin\in\binset_\node$, estimate (smoothed) $\quant_\node(\bin)\approx\frac{n^{(1)}_{\node,\bin}+0.5}{n^{(0)}_{\node,\bin}+n^{(1)}_{\node,\bin}+1}$
    \FOR{candidate split $(j,\tau)$}
        \STATE Split $\node\to(\child_l,\child_r)$ using feature $X^j$ at threshold $\tau$
        \FOR{$\child\in\{\child_l,\child_r\}$}
            \STATE $\Delta_{\node\to\child}(\bin)\gets \logit(\quant_\node(\bin))-\logit(\quant_\child(\bin))$ for shared bins $\bin \in \binset_\child$
            \STATE $\sigma_{\binset_\child}^2\left(\Delta_{\node\to\child}(\bin)\right) \gets \frac{\sum\limits_{\bin \in \binset_\child} |\bin| \left(\Delta_{\node\to\child}(\bin) - \frac{\sum\limits_{\bin' \in \binset_\child}|\bin'|\Delta_{\node\to\child}(\bin')}{\sum\limits_{\bin' \in \binset_\child}|\bin'|}\right)^2}{\sum\limits_{\bin \in \binset_\child} |\bin|}$
        \ENDFOR
        \STATE $\mathcal{L}(X^j,\tau;\node)\gets \frac{|\child_l|}{|\node|}\sigma_{\binset_{\child_l}}^2\left(\Delta_{\node\to\child_l}(\bin)\right)+\frac{|\child_r|}{|\node|}\sigma_{\binset_{\child_r}}^2\left(\Delta_{\node\to\child_r}(\bin)\right)$
    \ENDFOR
    \STATE Choose $(j^\star,\tau^\star)\in\arg\min_{(j,\tau)}\mathcal{L}(X^j,\tau;\node)$
    \IF{$\mathcal{L}(X^{j^\star},\tau^\star;m)\leq 0.999\sigma^2_{\mathcal B_m}(q)$ (0.1\% improvement)} 
        \STATE Split $\node$ on $X^{j^\star}$ at $\tau^\star$
    \ELSE 
        \STATE Stop at leaf node $\node$
    \ENDIF
\ENDFOR
\FOR{$k=1,\dots,K$}
    \STATE $\traindata_k \gets \{(x_i,0): i\in\majset,\ x_i\in\strat_k\}\ \cup\ \{(x_i,1): i\in\minset\}$
    \STATE Fit $\classifier_k \leftarrow \textsc{TRAIN}(\classifier,\traindata_k)$
    \STATE Weight $\alpha_k \leftarrow \mathrm{Perf}(\classifier_k)$ on validation
\ENDFOR
\RETURN $\hat p(x)=\frac{\sum_{k=1}^K \alpha_k\,\prob_{\classifier_k}(Y=1\mid X=x)}{\sum_{k=1}^K \alpha_k}$
\end{algorithmic}
\end{algorithm}

\newpage

\section{Additional Numerical Results}\label{app:res}

We provide time and peak memory usage in \ref{sec:timmem}, performance under varying levels of LMI in \ref{sec:res:vary}, class-specific performance and additional performance metrics in \ref{sec:res:othermet}, disaggregated results by classifier and dataset in \ref{sec:res:detail}, ablation tests in \ref{sec:res:abl}, hyperparameter sensitivity in \ref{sec:res:sens}, performance under alternative LMI mechanisms in \ref{sec:res:broaderLMI}, and performance under general class-count imbalance in \ref{sec:res:mar}.

\subsection{Time and Peak Memory Usage}\label{sec:timmem}

\begin{figure*}
\centering
\includegraphics[scale=0.3]{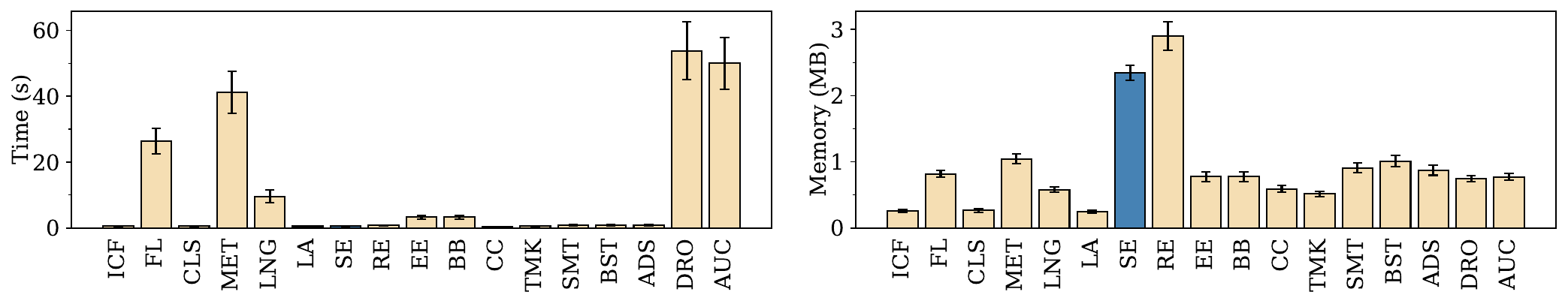}
\caption{Average computational time and peak memory usage.}
\label{fig:time}
\end{figure*}

Figure~\ref{fig:time} gives the average computational time and peak memory usage. Our algorithm is considerably faster than or competitive with existing methods, while using on average 1.5 MB higher memory than most other algorithms. These results provide good evidence that our method is scalable to larger datasets. 

\subsection{Varying Degrees of LMI}\label{sec:res:vary}

\begin{figure*}
\centering
\includegraphics[scale = 0.5]{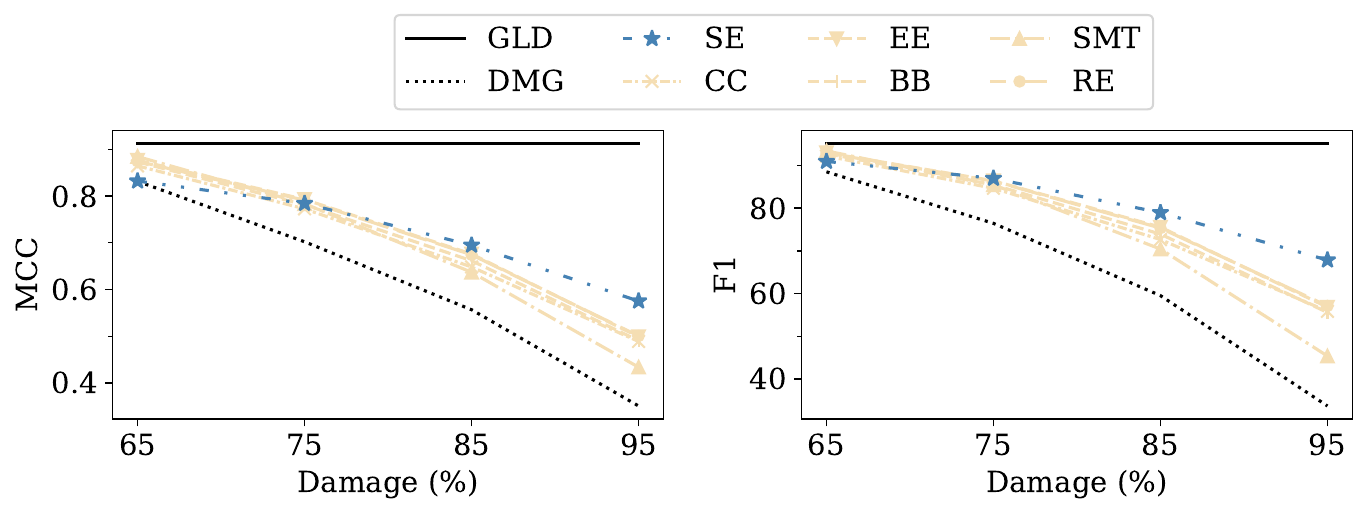}
\caption{MCC and F1 performance under varying levels of LMI.}
\label{fig:vary}
\end{figure*}

Figure~\ref{fig:vary} shows how MCC and F1 change with increasing LMI for the top-performing methods of Figure~\ref{fig:main}, averaged across datasets and classifiers. SE performs best at higher LMI levels (beyond 80\% or a 1:5 IR), with its advantage growing as imbalance increases. This is because lower LMI levels retain more of the underrepresented minority instances, and are closer to class-count imbalance scenarios. Note that at higher deletion rates, some feature regions may lose nearly all observed minority instances, making it impossible for any algorithm to recover performance.

\subsection{Class-specific and Additional Metrics}\label{sec:res:othermet}

\begin{figure*}
\centering
\includegraphics[scale = 0.265]{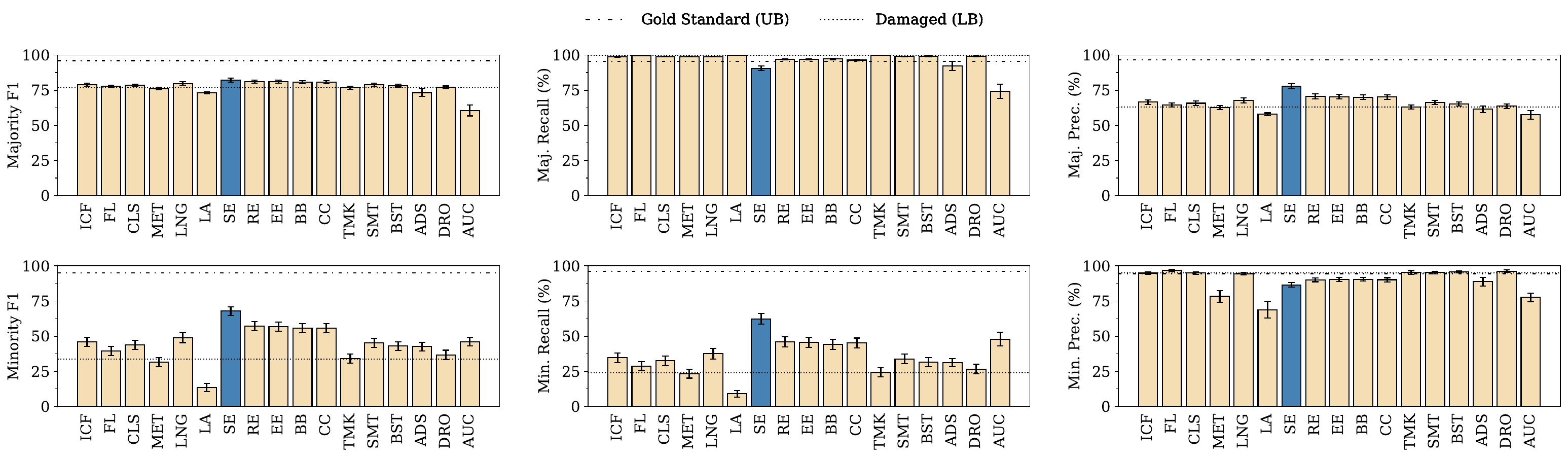}
\caption{Class-specific F1, Precision, and Recall under LMI.}
\label{fig:classwise}
\end{figure*}

To better evaluate the performance of SE, we report precision, recall, and F1 separately for both classes in Figure~\ref{fig:classwise}. SE achieves higher precision and F1 for the majority class, and higher recall and F1 for the minority class compared to all other competing methods. The results therefore indicate that SE achieves a more balanced allocation of predictive performance across classes under LMI, rather than improving minority detection at the cost of majority-class detection.

\begin{figure*}
\centering
\includegraphics[scale = 0.265]{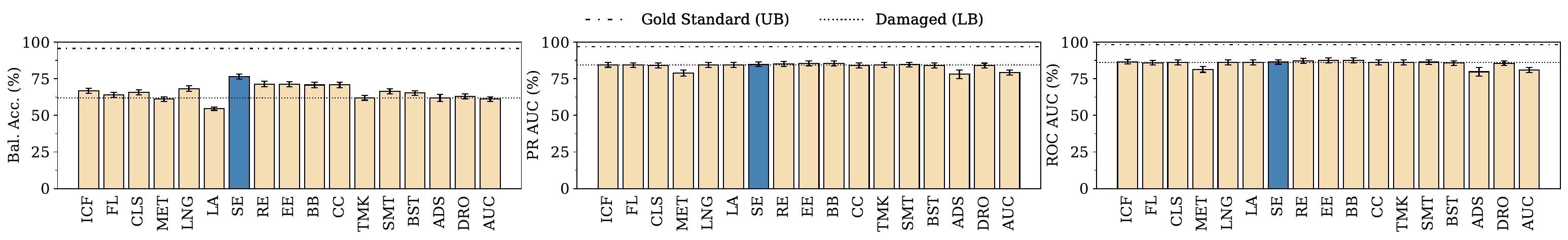}
\caption{Balanced accuracy, PR AUC, and ROC AUC performance under LMI.}
\label{fig:other}
\end{figure*} 

Figure~\ref{fig:other} gives the balanced accuracy, PR AUC, and ROC AUC performance, averaged over all datasets and classifiers. SE remains the best-performing method under balanced accuracy, but all methods perform similarly under AUC measures.  This shows that AUC metrics are much less informative under LMI, and confirms recent literature that has cautioned against treating AUC measures as automatically preferable under class imbalance \citep{mcdermott2024closer,richardson2024receiver}.

\subsection{Disaggregated Results by Classifier and Dataset}\label{sec:res:detail}

Table~\ref{tab:classifier_avg} gives the average MCC and F1 performance by classifier. SE achieves higher average MCC and F1 than the benchmarks for each of the five base classifiers.
\begin{table}
    \centering
    \caption{Detailed Performance by Base Classifier.}
    \label{tab:classifier_avg}
    \addtolength{\tabcolsep}{-0.3em}
\resizebox{\textwidth}{!}{
\begin{tabular}{llccccccccccccccccccc}
\toprule
\multirow{2}{*}{\textbf{Metric}} & \multirow{2}{*}{\textbf{Classifier}} & \textbf{UB} & \multicolumn{5}{c}{\textbf{Cost-based}} & \textbf{Debias} & \multicolumn{4}{c}{\textbf{Ensemble-based}} & \multicolumn{2}{c}{\textbf{Under-sampling}} & \multicolumn{3}{c}{\textbf{Over-sampling}} & \multicolumn{2}{c}{\textbf{Optimization}} & \textbf{LB} \\
\cmidrule(lr){3-3}
\cmidrule(lr){4-8}
\cmidrule(lr){9-9}
\cmidrule(lr){10-13}
\cmidrule(lr){14-15}
\cmidrule(lr){16-18}
\cmidrule(lr){19-20}
\cmidrule(lr){21-21}
 &  & GLD & ICF & FL & CLS & MET & LNG & LA & SE & RE & EE & BB & CC & TMK & SMT & BST & ADS & DRO & AUC & DMG \\
\midrule
\multirow{6}{*}{MCC} & LR & 0.81 & 0.54 & 0.45 & 0.51 & 0.28 & 0.49 & 0.12 & \textbf{0.57} & 0.52 & 0.52 & 0.52 & 0.53 & 0.34 & 0.52 & 0.49 & 0.51 & 0.38 & 0.42 & 0.34 \\
 & SVM & 0.90 & 0.38 & 0.37 & 0.36 & 0.33 & 0.42 & 0.22 & \textbf{0.53} & 0.47 & 0.47 & 0.46 & 0.45 & 0.36 & 0.42 & 0.40 & 0.39 & 0.36 & 0.33 & 0.35 \\
 & RF & 0.98 & 0.34 & 0.35 & 0.34 & 0.35 & 0.47 & 0.20 & \textbf{0.59} & 0.46 & 0.45 & 0.43 & 0.46 & 0.36 & 0.38 & 0.37 & 0.34 & 0.36 & 0.35 & 0.35 \\
 & NN & 0.91 & 0.51 & 0.42 & 0.49 & 0.28 & 0.50 & 0.08 & \textbf{0.59} & 0.56 & 0.56 & 0.55 & 0.54 & 0.37 & 0.46 & 0.44 & 0.41 & 0.39 & 0.30 & 0.36 \\
 & CAT & 0.96 & 0.41 & 0.38 & 0.39 & 0.29 & 0.42 & 0.19 & \textbf{0.59} & 0.49 & 0.49 & 0.49 & 0.46 & 0.34 & 0.39 & 0.38 & 0.36 & 0.37 & 0.07 & 0.34 \\
\cmidrule(lr){2-21}
 & \textbf{Avg} & 0.91 & 0.44 & 0.39 & 0.42 & 0.31 & 0.46 & 0.16 & \textbf{0.58} & 0.50 & 0.50 & 0.49 & 0.49 & 0.35 & 0.43 & 0.42 & 0.40 & 0.37 & 0.29 & 0.35 \\
\midrule
\multirow{6}{*}{F1} & LR & 89.6 & 61.5 & 48.3 & 58.2 & 32.6 & 54.2 & 10.4 & \textbf{70.2} & 61.5 & 61.6 & 60.7 & 61.6 & 34.2 & 59.3 & 55.2 & 59.1 & 40.3 & 56.0 & 33.8 \\
 & SVM & 94.4 & 38.2 & 36.1 & 35.6 & 32.2 & 44.2 & 19.7 & \textbf{62.5} & 53.8 & 53.7 & 52.6 & 52.3 & 34.3 & 43.9 & 41.4 & 41.3 & 34.6 & 32.4 & 33.9 \\
 & RF & 98.8 & 31.2 & 32.4 & 31.4 & 32.4 & 48.9 & 14.7 & \textbf{66.1} & 50.6 & 49.8 & 46.8 & 51.2 & 32.9 & 36.0 & 34.6 & 32.1 & 32.9 & 32.3 & 32.6 \\
 & NN & 95.3 & 56.9 & 43.8 & 54.4 & 30.6 & 55.0 & 6.9 & \textbf{71.3} & 64.0 & 63.8 & 62.7 & 61.2 & 36.7 & 48.6 & 46.5 & 44.2 & 40.0 & 49.2 & 36.1 \\
 & CAT & 97.9 & 42.2 & 36.5 & 39.3 & 29.9 & 42.2 & 15.5 & \textbf{69.1} & 55.6 & 55.1 & 55.1 & 52.8 & 32.2 & 38.9 & 37.6 & 36.1 & 35.6 & 60.4 & 31.8 \\
\cmidrule(lr){2-21}
 & \textbf{Avg} & 95.2 & 46.0 & 39.4 & 43.7 & 31.5 & 48.9 & 13.4 & \textbf{67.8} & 57.1 & 56.8 & 55.6 & 55.8 & 34.0 & 45.4 & 43.0 & 42.5 & 36.7 & 46.1 & 33.6 \\
\bottomrule
\end{tabular}}
\end{table}

Table~\ref{tab:dataset_avg} gives the average MCC and F1 performance by dataset. Our proposed SE approach gives higher F1 performance in five, and higher MCC performance in six out of eight datasets. Even when SE does not achieve the highest performance, it remains close to the highest-performing competing method. 

\begin{table}
    \centering
    \caption{Detailed Performance by Dataset.}
    \label{tab:dataset_avg}
    \addtolength{\tabcolsep}{-0.3em}
\resizebox{\textwidth}{!}{
\begin{tabular}{llccccccccccccccccccc}
\toprule
\multirow{2}{*}{\textbf{Metric}} & \multirow{2}{*}{\textbf{Dataset}} & \textbf{UB} & \multicolumn{5}{c}{\textbf{Cost-based}} & \textbf{Debias} & \multicolumn{4}{c}{\textbf{Ensemble-based}} & \multicolumn{2}{c}{\textbf{Under-sampling}} & \multicolumn{3}{c}{\textbf{Over-sampling}} & \multicolumn{2}{c}{\textbf{Optimization}} & \textbf{LB} \\
\cmidrule(lr){3-3}
\cmidrule(lr){4-8}
\cmidrule(lr){9-9}
\cmidrule(lr){10-13}
\cmidrule(lr){14-15}
\cmidrule(lr){16-18}
\cmidrule(lr){19-20}
\cmidrule(lr){21-21}
 &  & GLD & ICF & FL & CLS & MET & LNG & LA & SE & RE & EE & BB & CC & TMK & SMT & BST & ADS & DRO & AUC & DMG \\
\midrule
\multirow{9}{*}{MCC} & page-blocks0 & 0.89 & 0.35 & 0.32 & 0.33 & 0.29 & 0.36 & 0.20 & \textbf{0.40} & 0.35 & 0.35 & 0.35 & 0.32 & 0.30 & 0.35 & 0.34 & 0.38 & 0.31 & 0.29 & 0.29 \\
 & segment0 & 0.99 & 0.60 & 0.60 & 0.56 & 0.46 & \textbf{0.67} & 0.26 & 0.60 & 0.56 & 0.56 & 0.54 & 0.54 & 0.55 & 0.61 & 0.56 & 0.63 & 0.60 & 0.48 & 0.55 \\
 & abalone19 & 0.82 & 0.25 & 0.24 & 0.24 & 0.17 & 0.23 & 0.08 & 0.24 & 0.21 & 0.22 & 0.21 & 0.21 & 0.20 & 0.28 & 0.27 & \textbf{0.32} & 0.23 & 0.21 & 0.20 \\
 & shuttle-c0-vs-c4 & 1.00 & 0.82 & 0.78 & 0.80 & 0.76 & \textbf{0.92} & 0.51 & \textbf{0.92} & 0.89 & 0.90 & 0.90 & 0.87 & 0.77 & 0.85 & 0.78 & 0.42 & 0.77 & 0.41 & 0.77 \\
 & yeast3 & 0.87 & 0.34 & 0.29 & 0.32 & 0.10 & 0.36 & 0.03 & \textbf{0.63} & 0.43 & 0.42 & 0.40 & 0.50 & 0.21 & 0.32 & 0.32 & 0.33 & 0.22 & 0.22 & 0.21 \\
 & yeast4 & 0.85 & 0.31 & 0.27 & 0.30 & 0.21 & 0.31 & 0.06 & \textbf{0.48} & 0.38 & 0.39 & 0.38 & 0.35 & 0.22 & 0.31 & 0.30 & 0.33 & 0.25 & 0.23 & 0.22 \\
 & yeast5 & 0.97 & 0.43 & 0.32 & 0.41 & 0.28 & 0.41 & 0.11 & \textbf{0.72} & 0.64 & 0.62 & 0.61 & 0.58 & 0.29 & 0.39 & 0.37 & 0.41 & 0.30 & 0.27 & 0.28 \\
 & yeast6 & 0.91 & 0.39 & 0.33 & 0.38 & 0.19 & 0.42 & 0.06 & \textbf{0.62} & 0.55 & 0.54 & 0.54 & 0.54 & 0.29 & 0.37 & 0.38 & 0.38 & 0.30 & 0.26 & 0.29 \\
\cmidrule(lr){2-21}
 & \textbf{Avg} & 0.91 & 0.44 & 0.39 & 0.42 & 0.31 & 0.46 & 0.16 & \textbf{0.58} & 0.50 & 0.50 & 0.49 & 0.49 & 0.35 & 0.43 & 0.42 & 0.40 & 0.37 & 0.29 & 0.35 \\
\midrule
\multirow{9}{*}{F1} & page-blocks0 & 94.0 & 34.3 & 30.9 & 32.4 & 26.1 & 36.4 & 13.2 & 42.6 & 36.1 & 36.7 & 35.4 & 30.2 & 26.8 & 35.1 & 33.5 & 38.9 & 28.6 & \textbf{42.9} & 26.2 \\
 & segment0 & 99.4 & 65.1 & 65.5 & 61.1 & 49.1 & \textbf{72.4} & 23.8 & 71.2 & 62.7 & 62.5 & 61.0 & 63.4 & 59.1 & 66.1 & 61.1 & 67.8 & 65.0 & 63.2 & 59.1 \\
 & abalone19 & 89.9 & 26.0 & 21.3 & 24.1 & 17.0 & 22.6 & 4.6 & \textbf{37.0} & 28.4 & 28.8 & 28.1 & 26.1 & 16.1 & 28.9 & 27.5 & 35.2 & 18.8 & 29.3 & 15.8 \\
 & shuttle-c0-vs-c4 & 99.9 & 84.2 & 80.9 & 82.5 & 78.8 & \textbf{93.6} & 52.7 & 93.5 & 90.9 & 91.8 & 91.7 & 89.3 & 79.2 & 86.4 & 80.1 & 42.3 & 79.5 & 71.0 & 79.2 \\
 & yeast3 & 93.0 & 37.2 & 29.2 & 34.6 & 12.2 & 41.4 & 1.3 & \textbf{79.5} & 55.6 & 54.1 & 51.0 & 62.7 & 19.3 & 34.9 & 34.8 & 36.5 & 21.4 & 36.3 & 18.5 \\
 & yeast4 & 91.8 & 32.8 & 24.4 & 30.6 & 22.2 & 32.1 & 2.6 & \textbf{64.1} & 46.8 & 47.3 & 45.3 & 43.2 & 17.6 & 31.9 & 29.6 & 35.3 & 21.8 & 38.1 & 17.3 \\
 & yeast5 & 98.3 & 45.4 & 30.4 & 43.5 & 26.7 & 45.5 & 6.1 & \textbf{79.7} & 71.6 & 69.4 & 68.6 & 66.3 & 26.5 & 40.6 & 38.3 & 43.5 & 29.3 & 42.1 & 25.7 \\
 & yeast6 & 95.1 & 43.1 & 32.8 & 41.1 & 20.2 & 47.3 & 3.2 & \textbf{75.2} & 64.6 & 63.8 & 63.5 & 65.5 & 27.7 & 38.9 & 39.6 & 40.9 & 29.2 & 45.5 & 27.4 \\
\cmidrule(lr){2-21}
 & \textbf{Avg} & 95.2 & 46.0 & 39.4 & 43.7 & 31.5 & 48.9 & 13.4 & \textbf{67.8} & 57.1 & 56.8 & 55.6 & 55.8 & 34.0 & 45.4 & 43.0 & 42.5 & 36.7 & 46.1 & 33.6 \\
\bottomrule
\end{tabular}}
\end{table}

\subsection{Ablation Tests}\label{sec:res:abl}

\begin{figure}
     \centering
     \subfloat[][\centering Local vs Global (original setting) pairing of minority instances in SE.]{
         \centering
         \includegraphics[scale=0.4]{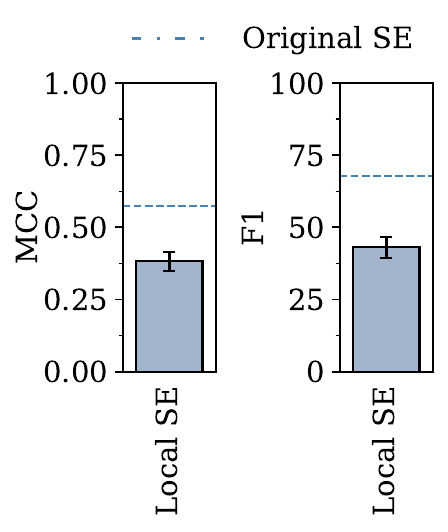}
         \label{fig:abl_pair}
     }\hspace{2mm}
     \subfloat[][\centering Uniform and F1-based vs MCC-based (original setting) weighting.]{
         \centering
         \includegraphics[scale=0.4]{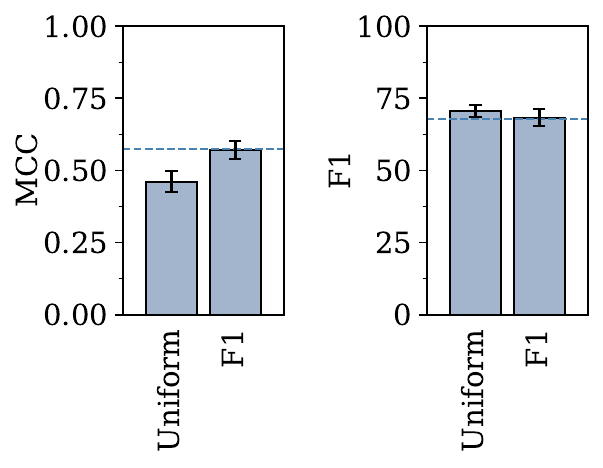}
         \label{fig:abl_uni}
     }\hspace{2mm}
     \subfloat[][\centering Alternative logistic regression (LR), CatBoost (CatB), and Random (RND) calculation of ranking scores $\eta(x)$, vs difference-in-means (original setting).]{
         \centering
         \includegraphics[scale=0.4]{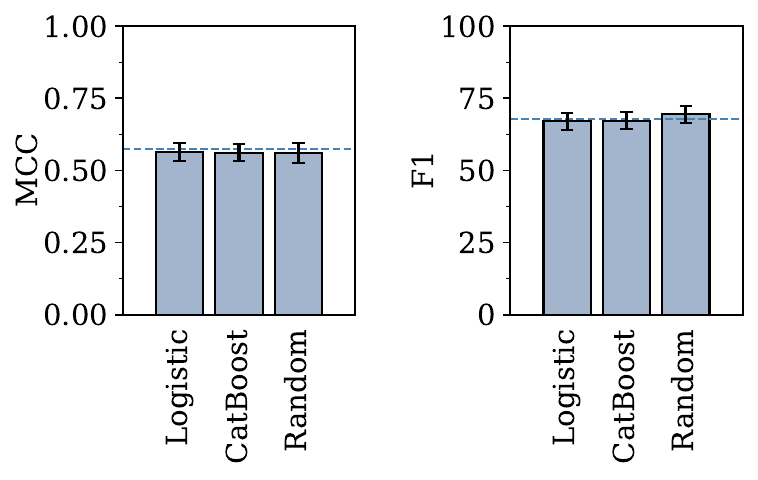}
         \label{fig:abl_prox}
     }
    \caption{Ablation Studies.}
    \label{fig:abl}
\end{figure}

We provide three sets of ablation tests in Figure~\ref{fig:abl}. First, we examine our bias-variance tradeoff strategy of combining all minority instances with the majority instances of each stratum $\strat_k \in \strata$. We compare this global minority pooling strategy with a local alternative that combines only the minority and majority instances within each stratum. As shown in Figure~\ref{fig:abl_pair}, global pooling achieves superior performance, supporting its role in mitigating the sparsity of minority observations within individual strata, and ultimately providing a better bias-variance tradeoff.

Second, we examine the contribution of MCC-based ensemble weighting by comparing it with uniform and F1-based averaging. As shown in Figure~\ref{fig:abl}(b), uniform averaging produces approximately six percentage points lower MCC, while improving F1 by approximately two points. F1-based averaging results in almost identical results to MCC-based averaging. These results suggest that performance-based weighting substantially improves results, despite relying on validation scores measured under the biased observation distribution.

Finally, we examine the sensitivity of SE to the choice of ranking proxy $\eta(x)$ used to construct the strata. Our original SE algorithm employs an inexpensive, model-agnostic difference-in-means direction, setting $\prox(x) \varpropto w^\top x$, where $w_j = (\mu_{1j} - \mu_{0j}) / \sqrt{(\sigma_{1j}^2 + \sigma_{0j}^2)/2}$, and $\mu_{yj}$ and $\sigma^2_{yj}$ denote the class-conditional mean and variance of feature $X^j \in X$ for class $Y=y$. We replace this proxy with confidence scores from logistic regression and CatBoost classifiers trained on $\odataset$, as well as a completely random ordering containing no predictive information about class membership. Results in Figure~\ref{fig:abl_prox} demonstrate robustness to model-based proxies, while random ordering preserves approximately the same F1 and reduces average MCC by approximately 0.03. 

Together, these findings indicate that SE benefits from global minority pooling and performance-based aggregation while remaining relatively insensitive to the specific ranking proxy used to construct its strata.

\subsection{Hyperparameter Sensitivity}\label{sec:res:sens}

We evaluate sensitivity to parameters of SE by varying the number of root node bins \(|\binset_0|\) from 50 to 150 (with 100 being the default), and the minimum leaf-size constraints from \(0.5\times\) to \(1.5\times\) their default values. Across these settings, average MCC and F1 vary by less than one percentage point, including on the smaller datasets. These results indicate that SE is not overly sensitive to precise choices of $|\binset_0|$ or the minimum leaf sizes.

\begin{figure}
     \centering
     \subfloat[][\centering Sensitivity to bin count of root node $|\binset_0|$.]{
         \centering
         \includegraphics[scale=0.48]{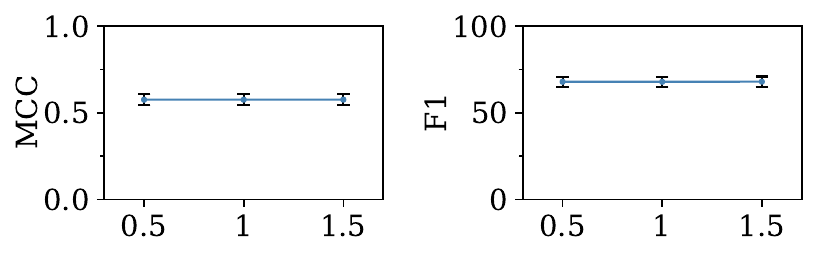}
         \label{fig:sens_bin}
     }
     \subfloat[][\centering Sensitivity to minimum leaf size $n_{\min}$.]{
         \centering
         \includegraphics[scale=0.48]{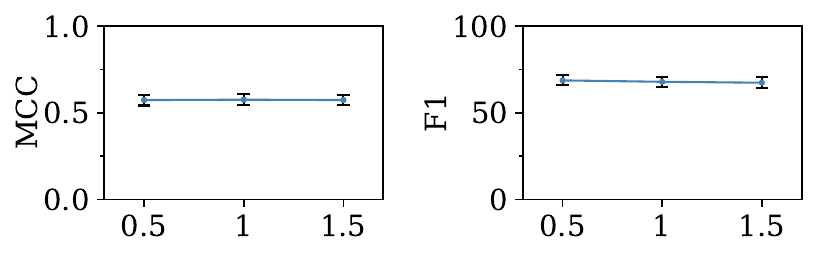}
         \label{fig:sens_min}
     }
    \caption{Sensitivity Results.}
    \label{fig:sensitivity}
\end{figure}

\subsection{Alternative LMI Mechanisms}\label{sec:res:broaderLMI}

Our primary experiments use a monotonic LMI mechanism driven by a single observed feature. We adopt this setting deliberately as a controlled benchmark rather than as a claim that real-world LMI is generally monotonic or uni-variate. A single-feature monotonic mechanism allows the source and severity of observation distortion to be varied transparently and consistently across datasets, while holding other aspects of the data-generating process fixed, making it possible to isolate the effect of increasing LMI severity and to compare methods under a common, interpretable distortion process. To assess whether the performance of SE is limited to this simplifying benchmark, we additionally evaluate five qualitatively different LMI mechanisms, including feature interactions, an unobserved driver, majority-concentration bias, non-monotonic deletion, and categorical or grouped deletion (see Appendix \ref{app:algos:deletion} for detailed definitions and overall simulation algorithm). 

Table~\ref{tab:broaderLMI} gives the average results for each mechanism. 
\begin{table}
    \centering
    \caption{Performance on Alternative LMI Mechanisms.}
    \label{tab:broaderLMI}
    \addtolength{\tabcolsep}{-0.3em}
\resizebox{\textwidth}{!}{
\begin{tabular}{llccccccccccccccccccc}
\toprule
\multirow{2}{*}{\textbf{Metric}} & \multirow{2}{*}{\textbf{LMI Mechanism}} & \textbf{UB} & \multicolumn{5}{c}{\textbf{Cost-based}} & \textbf{Debias} & \multicolumn{4}{c}{\textbf{Ensemble-based}} & \multicolumn{2}{c}{\textbf{Under-sampling}} & \multicolumn{3}{c}{\textbf{Over-sampling}} & \multicolumn{2}{c}{\textbf{Optimization}} & \textbf{LB} \\
\cmidrule(lr){3-3}
\cmidrule(lr){4-8}
\cmidrule(lr){9-9}
\cmidrule(lr){10-13}
\cmidrule(lr){14-15}
\cmidrule(lr){16-18}
\cmidrule(lr){19-20}
\cmidrule(lr){21-21}
 &  & GLD & ICF & FL & CLS & MET & LNG & LA & SE & RE & EE & BB & CC & TMK & SMT & BST & ADS & DRO & AUC & DMG \\
\midrule
\multirow{7}{*}{MCC} & Monotonic & 0.91 & 0.44 & 0.39 & 0.42 & 0.31 & 0.46 & 0.16 & \textbf{0.58} & 0.50 & 0.50 & 0.49 & 0.49 & 0.35 & 0.43 & 0.42 & 0.40 & 0.37 & 0.29 & 0.35 \\
 & Non-monotonic & 0.91 & 0.49 & 0.52 & 0.47 & 0.38 & 0.62 & 0.16 & \textbf{0.62} & 0.58 & 0.58 & 0.57 & 0.57 & 0.38 & 0.49 & 0.45 & 0.44 & 0.48 & 0.36 & 0.38 \\
 & Interaction & 0.91 & 0.47 & 0.41 & 0.45 & 0.32 & 0.49 & 0.16 & \textbf{0.58} & 0.54 & 0.54 & 0.53 & 0.53 & 0.37 & 0.46 & 0.43 & 0.43 & 0.39 & 0.31 & 0.36 \\
 & Unobserved & 0.90 & 0.53 & 0.46 & 0.52 & 0.34 & 0.55 & 0.17 & \textbf{0.62} & 0.59 & 0.60 & 0.59 & 0.60 & 0.40 & 0.55 & 0.51 & 0.51 & 0.41 & 0.34 & 0.40 \\
 & Majority-biased & 0.91 & 0.47 & 0.42 & 0.45 & 0.34 & 0.47 & 0.17 & \textbf{0.58} & 0.51 & 0.51 & 0.50 & 0.54 & 0.38 & 0.48 & 0.46 & 0.50 & 0.41 & 0.33 & 0.38 \\
 & Categorical & 0.91 & 0.52 & 0.53 & 0.50 & 0.37 & 0.60 & 0.18 & \textbf{0.62} & 0.60 & 0.59 & 0.58 & 0.57 & 0.40 & 0.53 & 0.51 & 0.52 & 0.50 & 0.39 & 0.39 \\
\cmidrule(lr){2-21}
 & \textbf{Avg} & 0.91 & 0.49 & 0.46 & 0.47 & 0.34 & 0.53 & 0.17 & \textbf{0.60} & 0.55 & 0.55 & 0.54 & 0.55 & 0.38 & 0.49 & 0.46 & 0.47 & 0.43 & 0.34 & 0.38 \\
\midrule
\multirow{7}{*}{F1} & Monotonic & 95.2 & 46.0 & 39.4 & 43.7 & 31.5 & 48.9 & 13.4 & \textbf{67.8} & 57.1 & 56.8 & 55.6 & 55.8 & 34.1 & 45.3 & 43.0 & 42.6 & 36.7 & 46.1 & 33.6 \\
 & Non-monotonic & 95.2 & 51.8 & 54.0 & 49.2 & 42.6 & 67.4 & 13.0 & \textbf{72.5} & 65.4 & 65.3 & 64.0 & 64.1 & 37.1 & 52.0 & 48.0 & 48.0 & 49.8 & 52.3 & 36.8 \\
 & Interaction & 95.2 & 49.1 & 41.4 & 46.9 & 33.2 & 51.8 & 13.1 & \textbf{68.2} & 61.2 & 60.9 & 59.7 & 59.3 & 35.8 & 47.7 & 44.3 & 45.0 & 38.5 & 47.5 & 35.4 \\
 & Unobserved & 94.2 & 57.7 & 47.5 & 55.5 & 37.2 & 59.5 & 14.4 & \textbf{73.6} & 70.3 & 70.8 & 70.0 & 67.9 & 40.9 & 59.6 & 55.8 & 56.5 & 42.7 & 50.3 & 40.3 \\
 & Majority-biased & 95.2 & 49.5 & 40.8 & 46.5 & 33.8 & 48.8 & 13.3 & \textbf{66.7} & 57.3 & 56.6 & 55.0 & 58.9 & 36.2 & 50.0 & 47.1 & 52.4 & 39.4 & 49.0 & 36.0 \\
 & Categorical & 95.2 & 56.3 & 56.4 & 54.0 & 42.5 & 65.0 & 15.3 & \textbf{74.3} & 67.9 & 67.5 & 66.3 & 65.6 & 40.4 & 57.2 & 55.1 & 56.6 & 52.9 & 55.0 & 39.6 \\
\cmidrule(lr){2-21}
 & \textbf{Avg} & 95.0 & 51.7 & 46.6 & 49.3 & 36.8 & 56.9 & 13.8 & \textbf{70.5} & 63.2 & 63.0 & 61.8 & 61.9 & 37.4 & 52.0 & 48.9 & 50.2 & 43.3 & 50.0 & 37.0 \\
\bottomrule
\end{tabular}}
\end{table}

SE remains the best-performing method under all six mechanisms. As expected, the advantage is smallest when the LMI-driving variable is unobserved. Let $X$ denote the observed features and $Z$ the unobserved driver. The effective observation propensity available through the observed data is

$$
\tilde{\pi}_1(x) = \mathbb{E}\!\left[\pi_1(X,Z)\mid X=x\right].
$$

When $Z$ is associated with observed features, SE can exploit the localized distortion expressed through those proxies. If the omitted driver is independent of $X$ and no observed feature carries information about the selection mechanism, then $\tilde{\pi}_1(x)$ becomes constant and the problem reduces to class-count imbalance with respect to the observed feature space. In that limiting case, no method using only $X$ can exploit localized structure. The positive gains in the omitted-driver setting therefore show that SE remains useful when the true driver is available only indirectly through correlated observed variables.

\subsection{Class-count Imbalance} \label{sec:res:mar}

\begin{figure*}
\centering
\includegraphics[scale = 0.3]{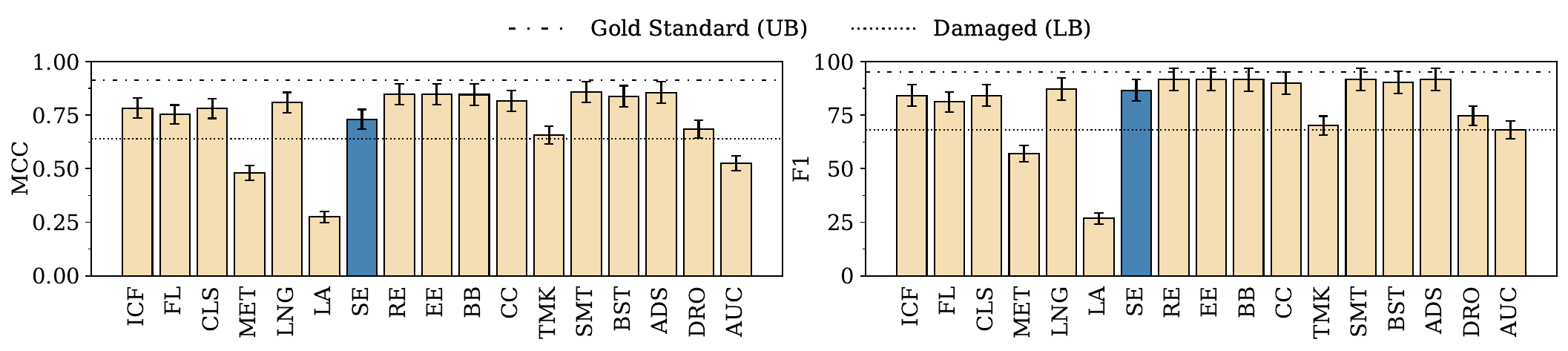}
\caption{Performance under class-count imbalance.}
\label{fig:mar}
\end{figure*} 

Figure~\ref{fig:mar} gives the average performance of all methods under the general class-count imbalance problem. Since the minority class retains sufficient representation under class-count imbalance, existing methods excel in mitigating imbalance. On the other hand, SE’s reliance on majority strata increases variance leading to slightly lower performance. 

Accordingly, SE should not be viewed as a universal replacement for standard class-imbalance methods. When no less-biased validation sample or external information about the observation process is available, the presence of LMI cannot be certified from $\odataset$ alone. In such settings, SE is most appropriately used as a robustness analysis when substantive knowledge makes localized observation bias plausible.

\section{Limitations}\label{app:limit}

Our study has several limitations. First, our method is developed and evaluated primarily for tabular classification. Extensions to non-tabular data may be possible by applying stratification to learned representations, although these adaptations are left for future work as they are best addressed using specialized algorithms and classifiers. Second, our split objective is not an exact identifier of the observation distortion $\gamma(x)$. As we prove in \S\ref{sec:theory}, this is impossible without external knowledge. Therefore, SE is a practical theory-motivated approximation. 

Finally, without additional assumptions or information, minority behavior cannot generally be identified in regions with no observed minority support. Likewise, LMI cannot be addressed from observed data alone when it is driven entirely by an unobserved feature that is uncorrelated with all observed features. This is because the observation distortion leaves no detectable localized structure in the observed feature space of such scenarios. Our approach is therefore most appropriate when LMI suppresses but does not entirely eliminate minority support in affected regions, and when the drivers of LMI are observed directly or are at least reflected through correlated observed features.

\end{document}